%% file: paper.tex
\documentclass[]{fairmeta}
\usepackage[disable]{todonotes}
\usepackage{enumitem}
\usepackage{tabularx} 
\usepackage{listings}
\usepackage{multirow}
\usepackage{makecell}
\usepackage{graphicx}
\usepackage{wrapfig}
\usepackage{tabularx}
\usepackage{textcomp}
\usepackage{placeins}
\usepackage{stfloats}
\usepackage{url}
\usepackage{verbatim}
\usepackage{titlesec}
\usepackage{tocloft}
\usepackage{adjustbox}
\usepackage{multirow}
\usepackage{pifont}
\usepackage{tikz}
\usepackage{comment}
\usepackage{amsmath,amssymb} % define this before the line numbering.
\usepackage{colortbl}  %彩色表格需要加载的宏包
\usepackage{color}
\usepackage{booktabs} 
\usepackage{hyperref}
\usepackage{graphicx}    % 用于插入图片
\usepackage{subcaption} 
\usepackage{multirow} % 表格多行合并
\usepackage{booktabs} % 表格更好的线条
\usepackage{subcaption} % 支持 subtable 和 subfigure
\RequirePackage{xspace}
\makeatletter
\DeclareRobustCommand\onedot{\futurelet\@let@token\@onedot}
\def\@onedot{\ifx\@let@token.\else.\null\fi\xspace}
\usepackage[most]{tcolorbox}
\usepackage{xcolor}
\usepackage{booktabs}
\usepackage{multirow}
\usepackage{makecell}
\usepackage{diagbox}
\usepackage{adjustbox}
\usepackage[table]{xcolor}
\usepackage{xltabular}
\setcitestyle{square,comma,numbers,sort&compress}

\makeatother

\definecolor{adptorange}{RGB}{248, 205, 172}
\definecolor{cmpblue}{RGB}{189, 215, 238}
\definecolor{cmpblue}{RGB}{189, 215, 238}
\definecolor{rynn}{RGB}{108,92,186} % 紫色边框近似

\definecolor{our_red}{RGB}{232,157,160}
\definecolor{our_blue}{RGB}{136,206,230}
\definecolor{our_orange}{RGB}{246,200,168}
\definecolor{our_green}{RGB}{178,211,164}

\definecolor{attn_code0}{RGB}{247,215,200}
\definecolor{attn_code1}{RGB}{238,169,139}
\definecolor{mlp_code0}{RGB}{204,201,221}
\definecolor{mlp_code1}{RGB}{102,95,153}

\definecolor{token_blue}{RGB}{84, 120, 140}
\definecolor{myblue}{RGB}{233, 241, 249}
\definecolor{mygray}{RGB}{99, 110, 114}
\definecolor{myred}{RGB}{255, 118, 117}
\definecolor{myyellow}{RGB}{255, 234, 167}
\definecolor{mygreen}{RGB}{216, 226, 204}
\definecolor{mypurple}{RGB}{162, 155, 254}
\definecolor{mybrown}{RGB}{215, 190, 154}
\definecolor{myorange}{RGB}{255, 220, 190}

\usepackage{tabularx} % 关键宏包：用于自适应宽度
\usepackage{pifont}       % \ding{xx}
\usepackage{bbding}       % \Checkmark,\XSolid,... (需要和pifont宏包共同使用)
\usepackage{fontawesome}
\usepackage{xspace}

\usepackage{float}

\newlength\savewidth

\newcolumntype{x}[1]{>{\centering\arraybackslash}p{#1pt}}
\newcolumntype{y}[1]{>{\raggedright\arraybackslash}p{#1pt}}
\newcolumntype{z}[1]{>{\raggedleft\arraybackslash}p{#1pt}}

\renewcommand{\paragraph}[1]{\vspace{1mm}\noindent\textbf{#1}}
\usepackage{colortbl}
\usepackage{xcolor}
\usepackage{wrapfig}
\usepackage{amssymb}

\renewcommand{\paragraph}[1]{\vspace{1.25mm}\noindent\textbf{#1}}

\usepackage{algorithm}
\usepackage{listings}

\definecolor{codeblue}{rgb}{0.25, 0.5, 0.5}
\definecolor{codekw}{rgb}{0.35, 0.35, 0.75}
\lstdefinestyle{Pytorch}{
    language = Python,
    backgroundcolor = \color{white},
    basicstyle = \fontsize{9pt}{8pt}\selectfont\ttfamily\bfseries,
    columns = fullflexible,
    aboveskip=1pt,
    belowskip=1pt,
    breaklines = true,
    captionpos = b,
    commentstyle = \color{codeblue},
    keywordstyle = \color{codekw},
}

\definecolor{green}{HTML}{009000}
\definecolor{red}{HTML}{ea4335}

\newcommand{\huggingface}{\raisebox{-1.5pt}{\includegraphics[height=1.05em]{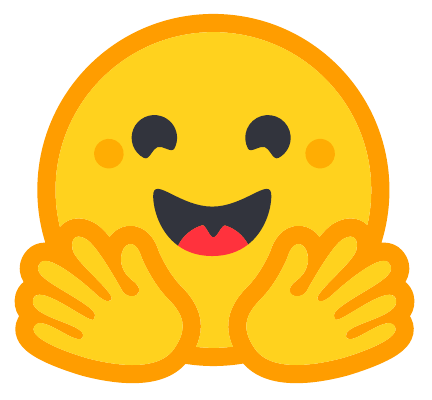}}\xspace}
\newcommand{\github}{\raisebox{-1.5pt}{\includegraphics[height=1.05em]{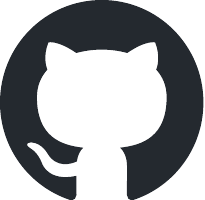}}\xspace}
\newcommand{\homepage}{\raisebox{-1.5pt}{\includegraphics[height=1.05em]{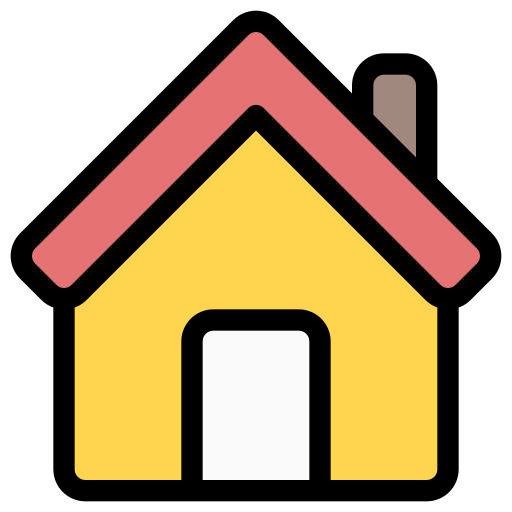}}\xspace}

\title{RynnUnify: Towards Unified Embodied Brain}

\author[*,\ddagger]{Minghao~Zhu}
\author[*,\ddagger]{Bohan~Hou}
\author[*]{Kun~Xiang}
\author[*]{Zixuan~Wang}
\author[*,\ddagger]{Yinzhou~Tang}
\author[*]{Yifei~Yang}
\author[*]{Zhikai~Wang}
\author[*]{Tianyi~Zhang}
\author[*]{Kehan~Li}

\author{Haoyu~Zhao}
\author{Siteng~Huang}
\author{Leigang~Qu}
\author{Sicong~Leng}
\author{Dongchi~Huang}
\author{Xiao~Lin}
\author{Ronghao~Dang}
\author{Mingxiu~Chen}
\author{Shijian~Lu}
\author{Yue~Wang}
\author{Xiaodan~Liang}
\author{Zhenyu~Zeng}
\author{Deli~Zhao}

\author[\dagger]{Chengju~Liu}
\author[\dagger]{Jianfei~Yang}
\author[\ddagger]{Xin~Li}

\affiliation{DAMO Academy, Alibaba Group}
\affiliation{Hupan Lab}

\contribution[*]{Core contributors}
\contribution[\ddagger]{Project leaders}
\contribution[\dagger]{Co-corresponding authors}
\input{sec/0_abstract}
\date{\today}

\begin{document}
\thispagestyle{firstheader}

% [original]
% \maketitle

% [revised]
\maketitle

%\pagestyle{empty}
%\setcounter{tocdepth}{3}

% [original]
% \par
% \begingroup
%     \setlength{\parskip}{0pt}
%     \sbox{\RynnTeaserCaption}{%
%         \begin{minipage}{\textwidth}
%             \captionsetup{font=small,skip=4pt}
%             \captionof{figure}{\textbf{Overview of RynnUnify.} The model connects embodied understanding and grounding, long-horizon visual planning, agentic generation-enhanced reasoning, and VLA control. Ground-truth video and AI-refined robot illustrations are marked in the figure.}
%             \label{fig:teaser}
%         \end{minipage}%
%     }
%     \setlength{\RynnTeaserImageHeight}{\dimexpr
%         \textheight-\pagetotal
%         -\ht\RynnTeaserCaption-\dp\RynnTeaserCaption
%         -2\baselineskip\relax}
%     \noindent\begin{minipage}{\textwidth}
%         \centering
%         \includegraphics[width=\linewidth,height=\RynnTeaserImageHeight,keepaspectratio]{figures/RynnUnify_Teaser_v15_cropped.pdf}
%         \par
%         \usebox{\RynnTeaserCaption}
%     \end{minipage}
%     \par
% \endgroup

% [revised]
\FloatBarrier
\vspace{-0.6em}
\begin{figure*}[!h]
    \centering
    \includegraphics[width=0.92\linewidth]{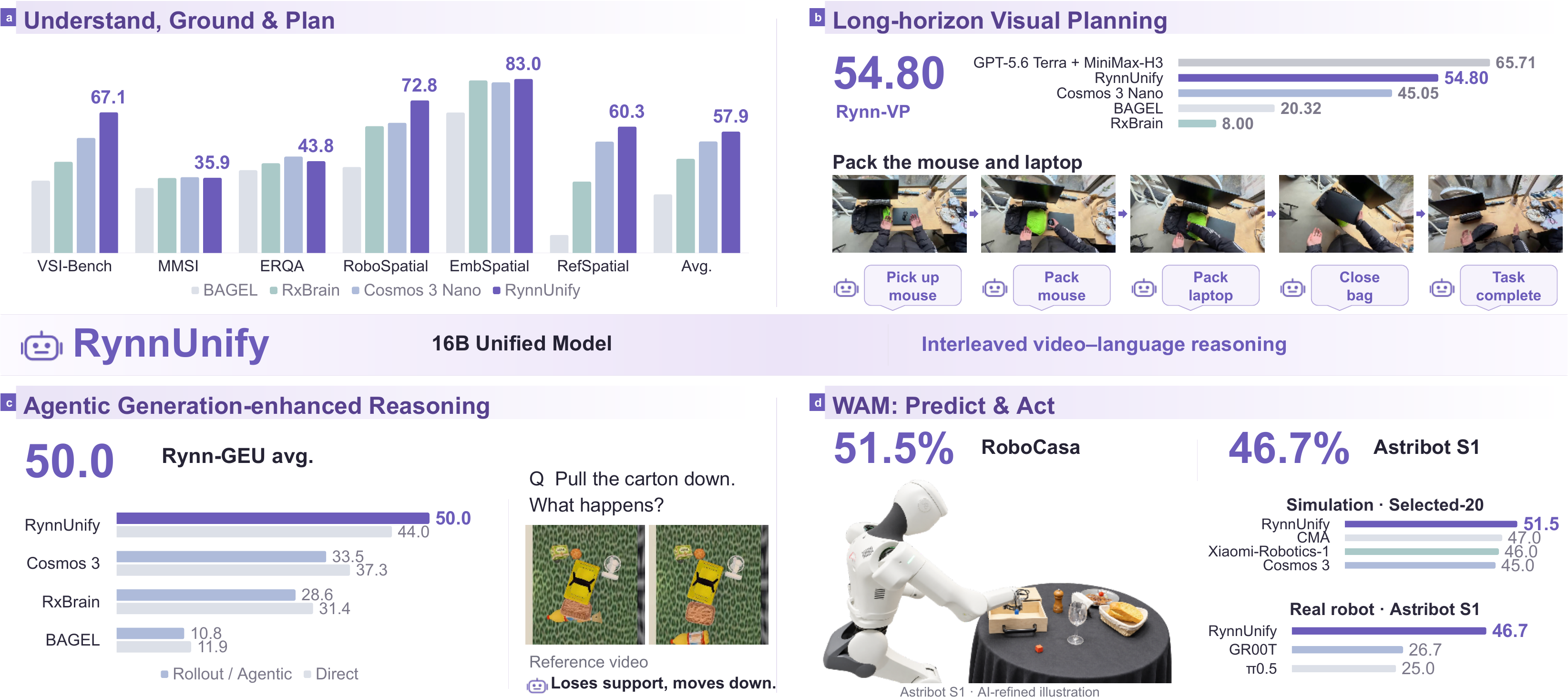}
    \caption{RynnUnify unifies embodied understanding, visual planning, agentic reasoning, and robot action.}
    \label{fig:teaser}
\end{figure*}
\FloatBarrier

\noindent
\input{sec/1_introduction}

\input{sec/2_model_architecture}
\input{sec/3_Data}
\input{sec/4_Training}

\input{sec/5_Specific_Task_Supervised_Finetuning}

\input{sec/6_Evaluation}
\input{sec/7_Conclusion}

\bibliographystyle{assets/plainnat}
\bibliography{paper}
\newpage
\beginappendix
\input{sec/appendix}
\end{document}

%% file: sec/0_abstract.tex
\abstract{
\vspace{-1em}
We introduce RynnUnify, a 16B unified embodied model that combines strong embodied understanding with long-horizon visual planning through interleaved video--language reasoning. With capabilities spanning spatial reasoning, physical prediction, and grounding, RynnUnify connects an understanding of the physical world with visual anticipation of how it evolves across successive subtasks. Building on these capabilities, we further develop an agentic extension through reinforcement learning to selectively use video predictions for reasoning about unobserved states and interaction outcomes. We also develop a world-action extension of the full RynnUnify model, translating its understanding and prediction capabilities into robot control. To evaluate these capabilities, we introduce Rynn-VP and Rynn-GEU for long-horizon visual planning and anticipatory reasoning. Experiments show that RynnUnify achieves the highest average embodied understanding score among the evaluated unified multimodal models and improves the Rynn-VP overall score by 9.75 points over Cosmos 3 Nano. Its agentic extension improves the Rynn-GEU average by 6.0 points over direct reasoning, while the world-action extension achieves 51.5\% success on 20 selected composite RoboCasa365 tasks and 46.7\% average success across three real-robot tasks. Together, these results establish RynnUnify as a shared foundation for embodied understanding, visual planning, and robot control, taking a step toward a unified embodied brain.

\vspace{-1em}
\begin{center}
    \renewcommand{\arraystretch}{1.2}
    \begin{tabular}{ll}
        \homepage & \url{https://alibaba-damo-academy.github.io/RynnUnify.github.io} \\
        \github & \url{https://github.com/alibaba-damo-academy/RynnUnify} \\
        \huggingface & \href{https://huggingface.co/collections/Alibaba-DAMO-Academy/rynnunify}{\nolinkurl{https://huggingface.co/collections/Alibaba-DAMO-Academy/RynnUnify}} \\
    \end{tabular}
    \vspace{-1em}
\end{center}
}

%% file: sec/1_introduction.tex
\section{Introduction}

Developing general robot brains~\cite{damo2026rynnbrain,li2026rynnbrain11capablegeneralizable} requires models that understand the physical world, anticipate interaction outcomes, and translate task intentions into action. Across interdependent operations, a robot must identify subgoals, predict scene changes, and determine how they affect subsequent decisions. These demands motivate unified embodied models that coordinate understanding, world prediction, and action generation~\cite{cosmos3,liang2026rxbrain}.

Existing unified multimodal models (UMMs) have made encouraging progress in jointly modeling understanding and generation~\cite{cosmos3,liang2026rxbrain,deng2025bagel}. Their coordination requires strong embodied understanding to ground task intentions, reason about physical interactions, and interpret predicted outcomes. Sustaining this interaction over long horizons remains challenging as scene changes and prediction errors accumulate, while static visual predictions can overlook intermediate dynamics. Addressing these challenges requires interleaved reasoning~\cite{hou2026interlv} that connects embodied understanding with temporally coherent video prediction.

To this end, we introduce RynnUnify, a 16B unified embodied model built upon the Mixture-of-Transformers (MoT) architecture~\cite{liang2025mixtureoftransformers} of Cosmos3~\cite{cosmos3}, comprising an 8B autoregressive reasoner and an 8B diffusion generator. Centered on \textbf{interleaved video-language reasoning}, RynnUnify combines strong embodied understanding with video prediction to support understanding-guided long-horizon visual planning: the reasoner guides successive video predictions, whose outcomes provide context for subsequent subtask planning. To explore generation-enhanced embodied understanding, we further post-train an additional 8B agent from RynnUnify's reasoner through reinforcement learning with final-answer feedback. The agent dynamically chooses between direct answering and invoking the frozen RynnUnify model to generate videos, which it then interprets to answer questions about unobserved future states and interaction outcomes. To bring the UMM's understanding and world-prediction capabilities to robot control, we also extend the full RynnUnify model to a World-Action Model (WAM) through latent planning and joint video-action training.

We perform extensive pretraining on general multimodal understanding, spatial reasoning, physical prediction, and affordance grounding, complemented by subtask planning, world-modeling descriptions, and question answering with intermediate video context. This supervision strengthens foundational embodied understanding while preparing the reasoner to formulate task-relevant visual predictions and interpret their outcomes. Generation training covers long-horizon egocentric activities, robot manipulation, and embodied reasoning videos. We introduce scene memory and self-forcing~\cite{huang2025selfforcing} to improve visual consistency and robustness during long-horizon generation.

Dedicated benchmarks remain lacking for assessing sustained visual planning and the contribution of video prediction to future-state reasoning. We therefore introduce Rynn-VP and Rynn-GEU. Rynn-VP evaluates whether visual plans remain consistent with the task and preceding operations over long horizons. Rynn-GEU focuses on action consequences, physical changes, and spatial relationships after motion or viewpoint changes, whose answers concern states not directly observable in the input. Its direct setting tests whether models can infer these states without explicit video generation, while its generation-assisted setting tests whether predicted videos improve answers to the same questions.

Extensive evaluations demonstrate state-of-the-art embodied understanding among the evaluated UMMs on public benchmarks, together with strong video generation performance. RynnUnify also demonstrates sustained planning on Rynn-VP and strong direct anticipatory reasoning on Rynn-GEU, with further GEU improvements through its agentic extension. For robot control, the WAM extension achieves a 51.5\% success rate on 20 selected composite RoboCasa365~\cite{robocasa365} tasks. Experiments on Astribot S1 further demonstrate effective multi-step manipulation involving body reorientation, precise object handling, and history-dependent execution.

Figure~\ref{fig:teaser} summarizes RynnUnify's capabilities and representative evaluation results.

Our main contributions are summarized as follows:
\begin{itemize}
    \item We introduce RynnUnify, a unified embodied model that combines strong embodied understanding with interleaved video-language reasoning for long-horizon visual planning.

    \item We post-train an agent from RynnUnify's reasoner to explore generation-enhanced embodied understanding, demonstrating the utility of video predictions for reasoning about unobserved future states.

    \item We extend the full RynnUnify UMM to a WAM through latent planning and joint video-action training, and validate its capabilities in simulated and real-robot manipulation.

    \item We introduce Rynn-VP and Rynn-GEU for long-horizon visual planning and anticipatory embodied reasoning, complemented by extensive evaluations on public benchmarks that establish RynnUnify's state-of-the-art embodied understanding among the compared UMMs.
\end{itemize}

%% file: sec/2_model_architecture.tex
\section{Model Architecture}
\label{sec:architecture}
\begin{figure*}[t]
    \centering
    \includegraphics[width=\textwidth]{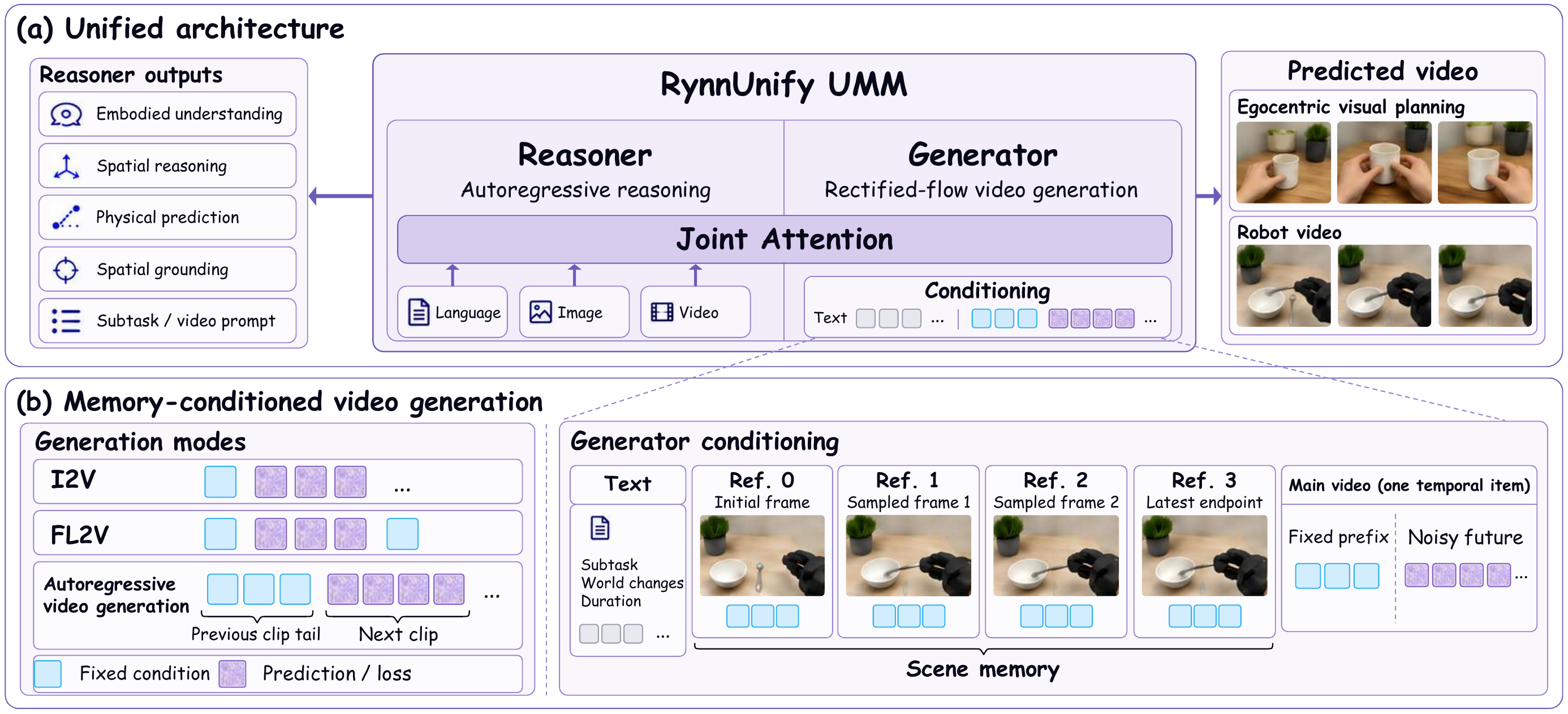}
    \caption{\textbf{Overview of the RynnUnify architecture.} (a) An autoregressive reasoner and a rectified-flow video generator interact through joint attention. (b) Memory-conditioned video generation supports I2V, FL2V, and autoregressive video continuation. Scene-memory frames provide persistent visual references, while a fixed temporal prefix carries recent motion and interaction context across consecutive clips.}
    \label{fig:architecture}
\end{figure*}
As shown in Figure~\ref{fig:architecture}, RynnUnify adopts the Mixture-of-Transformers (MoT)~\cite{liang2025mixtureoftransformers} architecture of Cosmos3~\cite{cosmos3}. The model comprises an autoregressive reasoner for multimodal understanding and language generation, and a diffusion generator for predicting continuous visual latents through rectified flow~\cite{liu2022flow}. The two branches use separate Transformer parameters and interact through joint attention, allowing the generator to attend to representations produced by the reasoner. We retain the reasoner architecture unchanged and extend the generator's visual conditioning to incorporate cross-chunk scene memory and support first-and-last-frame-conditioned generation. These extensions provide persistent scene references and continuous interaction context for visual predictions across successive subtasks.
\subsection{Reasoner}
\label{sec:reasoner}

We retain the autoregressive reasoner of Cosmos3~\cite{cosmos3} without architectural modifications. The reasoner processes multimodal observations and language context to support embodied understanding, spatial grounding, and task planning. For visual planning, it takes the task goal, initial and current observations, and preceding subtask history as input, and predicts the next executable subtask together with a detailed description of its anticipated visual evolution and an estimated duration. These outputs provide the generator with structured textual conditions specifying the intended interaction and its temporal extent. During multi-step rollout, the final frame of each generated clip becomes the updated observation for subsequent reasoning, connecting subtask planning with the visual consequences of preceding predictions.

\subsection{Visual tokenization and input organization.}
For video generation, we encode visual inputs using a causal video VAE with spatial and temporal compression factors of 16 and 4, respectively. The resulting latents are spatially patchified into visual tokens. Both scene-memory frames and the main video follow this VAE pathway and are processed by the generator branch.

Each scene-memory frame is encoded independently, producing a group of visual tokens that represents its spatial content. Four reference frames therefore correspond to four groups of tokens, whose individual lengths depend on the frame resolutions. When scene memory is enabled, these groups are placed after the reasoner's textual context and before the main video tokens:
\begin{equation}
    \mathbf{S}_{t}
    =
    \big[
    \mathbf{S}_{\mathrm{AR}};
    \underbrace{
        \mathbf{Z}_{\mathrm{mem}}^{(0)};
        \ldots;
        \mathbf{Z}_{\mathrm{mem}}^{(3)}
    }_{\mathbf{Z}_{\mathrm{mem}}};
    \mathbf{z}_{t}
    \big],
\end{equation}
where $\mathbf{S}_{\mathrm{AR}}$ denotes the textual conditioning tokens processed by the reasoner, $\mathbf{Z}_{\mathrm{mem}}^{(i)}$ denotes the visual tokens of the $i$-th reference frame, and $\mathbf{z}_{t}$ denotes the main video tokens at denoising time $t$. Both $\mathbf{Z}_{\mathrm{mem}}$ and $\mathbf{z}_{t}$ belong to the generator's visual sequence. The memory groups retain their boundaries as independently encoded visual items; their placement before the main video is a concatenation of token groups, rather than an insertion of additional frames into the main video's temporal axis. For generation tasks without scene memory, $\mathbf{Z}_{\mathrm{mem}}$ is omitted.

At each denoising step, the generator processes the memory and main video tokens through its existing attention layers while attending to the reasoner's textual representations. The main video can therefore access the scene references directly, without introducing a separate memory encoder or additional cross-attention modules.

\subsection{Cross-chunk scene memory.}
Long-horizon planning requires the model to retain visual information beyond the current observation, particularly when objects or scene regions leave the field of view. We maintain four reference frames comprising a persistent initial observation, two keyframes selected across earlier history, and a reference associated with the most recent step. This selection preserves both broader scene coverage and recent task context. When the history is insufficient, additional frames are selected from available historical clips; at the first step, the initial observation is repeated to fill the reference slots. During rollout, the initial observation remains fixed while the historical references are updated from generated videos.

We complement these sparse references with a continuous video prefix from the preceding chunk. Unlike the independently encoded memory frames, this prefix occupies the beginning of the main video's temporal sequence. The preceding tail and the current segment form a continuous sequence, with the shared boundary frame included only once. The reference frames serve as visual anchors for scene appearance and layout, while the temporal prefix preserves recent motion and interaction context. Together, they form cross-chunk scene memory for successive visual predictions.

\subsection{Generation modes and fixed conditions.}
We support three generation modes through different arrangements of visual conditions within the main video sequence. In image-to-video (I2V) generation, the first latent frame is fixed to the encoded starting observation, and the model predicts the subsequent video. In first-and-last-frame-to-video (FL2V) generation, both the first and final latent frames are fixed, providing boundary conditions for generating the intervening transition. In cross-chunk video continuation, the initial latent frames are fixed to a continuous prefix from the preceding video's tail, allowing the model to predict the next segment from recent motion and interaction context. During multi-step rollout, this prefix is extracted from the previously generated video.

Across these modes, the scene-memory latents and the conditioning latents within the main video remain fixed throughout denoising. The generator updates the remaining video positions while attending to the textual context and visual conditions. During training, only the positions to be predicted contribute to the generation loss. The same conditioning interface accommodates observed and self-generated visual inputs, supporting the self-forcing strategies used during generator fine-tuning.

\subsection{Textual and temporal conditioning.}
The generator receives structured textual conditions containing the subtask instruction, a detailed description of the anticipated visual evolution, and the requested duration. These conditions connect task-level reasoning with the content and temporal extent of each video prediction. We support variable-length video sequences of the VAE-compatible form $4n+1$ frames and retain the backbone's three-dimensional rotary position encoding and frame-rate modulation to represent spatial and temporal coordinates. For cross-chunk continuation, the requested duration describes the new operation and excludes the historical prefix. The reasoner can therefore specify what should happen next and for how long, while the generator predicts the corresponding visual changes using the scene memory and temporal conditions.

%% file: sec/3_Data.tex
\section{Data}

% Required packages: booktabs, tabularx, array, multirow
% Required packages: booktabs, tabularx, array, multirow
% Required packages: booktabs, tabularx, array, multirow
% Required packages: booktabs, tabularx, array, multirow
% Preamble
\lstdefinestyle{rynnformat}{
    basicstyle=\ttfamily\scriptsize\color{black},
    frame=single,
    framerule=0.4pt,
    framesep=4pt,
    rulecolor=\color{rynn},
    backgroundcolor=\color{rynn!4},
    breaklines=true,
    columns=fullflexible,
    keepspaces=true,
    showstringspaces=false,
    aboveskip=8pt,
    belowskip=8pt
}

Table~\ref{tab:training_data} summarizes the data used to fine-tune the reasoner and generator.

\subsection{General Understanding Data}
To maintain broad embodied understanding and reasoning capabilities, we use a 10\% subset of the RynnBrain pretraining corpus~\cite{li2026rynnbrain11capablegeneralizable} as the foundation of our understanding data. Reasoning about embodied tasks requires the model to interpret task intentions, recognize objects and their properties, understand spatial relationships, and track changes across successive observations. These capabilities are particularly important for long-horizon visual planning, where each subtask must be grounded in the current scene and considered in relation to subsequent operations. They also support the interaction between understanding and video generation: the model needs to identify which aspects of the scene are relevant to a task, formulate the corresponding visual prediction, and interpret the predicted changes to inform further reasoning.

The data cover general image and video understanding, visual question answering, temporal reasoning, and instruction following, together with embodied cognition tasks including object understanding, spatial reasoning, counting, and egocentric task understanding. Spatially grounded supervision further covers object, area, and affordance localization, trajectory prediction, and manipulation planning, connecting language instructions with explicit locations and interaction targets. We combine this foundational supervision with our visual planning and embodied perception and reasoning data to develop interleaved video--language reasoning while retaining a broad range of understanding capabilities. This combination is intended to support both direct responses to embodied questions and the interpretation of visual evidence produced during task-oriented video prediction.

\begin{table*}[t]
\centering
\caption{
Overview of the data used for reasoner and generator fine-tuning in RynnUnify, covering general understanding, visual planning, and embodied perception and reasoning. We report sample counts for understanding and clip counts and approximate durations for video generation.
}
\label{tab:training_data}

\begingroup
\footnotesize
\color{black}
\arrayrulecolor{black}
\setlength{\tabcolsep}{5pt}
\renewcommand{\arraystretch}{1.16}

\begin{tabularx}{\textwidth}{
    @{}
    >{\raggedright\arraybackslash}p{0.145\textwidth}
    >{\raggedright\arraybackslash}p{0.115\textwidth}
    >{\raggedright\arraybackslash}X
    r
    r
    @{}
}
\toprule
\makecell[cl]{\textbf{Category}}
& \makecell[cl]{\textbf{Subtask /}\\\textbf{Mode}}
& \makecell[cl]{\textbf{Data Sources}}
& \makecell[cr]{\textbf{Samples /}\\\textbf{Clips}}
& \makecell[cr]{\textbf{Duration}\\\textbf{(h)}} \\
\midrule

% ============================================================
% Understanding data
% ============================================================
\rowcolor{rynn!6}[0pt][0pt]
\multicolumn{5}{@{}l@{}}{\textbf{Understanding Data}} \\
\midrule

General Understanding
& General
& RynnBrain~\cite{damo2026rynnbrain}
& \makecell[r]{$\sim$2.0M\\(10\% subset)}
& -- \\

\midrule

\multirow[t]{3}{=}{Visual Planning}
& Human
& RynnBrain-Planning~\cite{damo2026rynnbrain}, HiFi-UMI~\cite{ai2026hifi}, OpenAoE~\cite{li2026open}
& 550,000
& -- \\

\addlinespace[2pt]
& Robot
& AgiBot-Beta~\cite{bu2025agibot_arxiv}, RoboMIND~\cite{wu2025robomind}, RoboCOIN~\cite{wu2025robocoin}, RoboTwin~\cite{Mu_2025_CVPR}, HIW-500~\cite{hiw500_2026}
& 600,000
& -- \\

\addlinespace[2pt]
&
& RoboCasa365~\cite{robocasa365}
& 86,800
& -- \\

\midrule

\multirow[t]{2}{=}{Embodied Perception and Reasoning}
& Reasoning
& RynnUnify-Spatial, RynnUnify-Phys,
  RynnUnify-ActionReasoning, RynnUnify-Mess
& 153,195
& -- \\

\addlinespace[2pt]
& Perception
& RynnUnify-Affordance, RynnUnify-Area,
  RynnUnify-Trajectory
& 43,405
& -- \\

\midrule

% ============================================================
% Video generation data
% ============================================================
\rowcolor{rynn!6}[0pt][0pt]
\multicolumn{5}{@{}l@{}}{\textbf{Video Generation Data}} \\
\midrule

\multirow[t]{2}{=}{Visual Planning}
& TI2V 
& RynnBrain-Planning~\cite{damo2026rynnbrain}, HiFi-UMI~\cite{ai2026hifi}, OpenAoE~\cite{li2026open}, HIW-500~\cite{hiw500_2026},
  AgiBot-Beta~\cite{bu2025agibot_arxiv}, RoboMIND~\cite{wu2025robomind}, RoboCOIN~\cite{wu2025robocoin}, RoboTwin~\cite{Mu_2025_CVPR}
& 806,685
& 966.3 \\

\addlinespace[2pt]
& TI2V 
& RoboCasa365~\cite{robocasa365}
& 80,258
& 146.4 \\

\midrule

\multirow[t]{2}{=}{Embodied Perception and Reasoning}
& FL2V
& RynnUnify-Spatial
& 117,921
& 146.0 \\

\addlinespace[2pt]
& TI2V
& RynnUnify-Spatial, RynnUnify-Phys,
  RynnUnify-ActionReasoning, RynnUnify-Mess,
  RynnUnify-Affordance, RynnUnify-Area,
  RynnUnify-Trajectory
& 78,594
& 111.4 \\

\midrule
\multicolumn{3}{@{}l}{\textbf{Total Video Generation Data}}
& \textbf{1,083,458}
& \textbf{1,370.1} \\
\bottomrule
\end{tabularx}

\vspace{3pt}
\begin{minipage}{\textwidth}
\footnotesize
\textit{Notes.}
TI2V: text-and-image-to-video generation;
FL2V: first-and-last-frame-conditioned video generation.
Shared planning clips are counted once, and durations exclude repeated training exposure.
\end{minipage}

\endgroup
\end{table*}

\subsection{Visual Planning Data}
Long-horizon visual planning requires the reasoner to determine successive subtasks from an evolving scene and the generator to predict the visual consequences of their execution. Developing these complementary capabilities requires supervision that connects task intentions, intermediate decisions, and continuous scene changes across multiple operations. We therefore curate human and robot manipulation demonstrations and organize them into temporally ordered sequences of subtask instructions and corresponding video clips, supporting the fine-tuning of both the reasoner and the generator.

\paragraph{Data collection and annotation.}
Our data sources include RynnBrain-Planning~\cite{damo2026rynnbrain}, HiFi-UMI~\cite{ai2026hifi}, OpenAoE~\cite{li2026open}, HIW-500, AgiBot-Beta~\cite{bu2025agibot_arxiv}, RoboMIND~\cite{wu2025robomind}, RoboCOIN~\cite{wu2025robocoin}, RoboTwin~\cite{Mu_2025_CVPR}, and RoboCasa365~\cite{robocasa365}, covering human and UMI demonstrations, real-robot interactions, and simulated manipulation. We additionally collect egocentric loco-manipulation videos in indoor environments as part of our RynnUnify data collection. Participants perform step-by-step object interactions using hand motions that imitate robotic manipulation, with each operation explicitly demonstrated in the recorded video. These recordings capture how successive operations change object states and spatial relationships throughout a task.

For datasets with explicit subtask annotations, we use the available instructions and temporal boundaries to associate each operation with its corresponding visual observations. For demonstrations containing extended multi-step activities but lacking explicit subtask decomposition, we use Qwen3.6-27B to perform temporal segmentation and relabeling. Each resulting segment is paired with a semantically meaningful subtask instruction describing the operation performed. We aim to keep video clips between 2 and 10 seconds whenever possible, yielding manageable training segments while preserving the semantic completeness of each operation and the temporal evolution of the interaction.
We then use GPT-5.6-Terra to rewrite each subtask instruction into a structured visual prompt grounded in the corresponding clip, describing the scene configuration, relevant objects, interaction progression, and resulting state changes.
These prompts supervise the reasoner's visual-prompt formulation and serve as textual conditions for training the generator on the corresponding video clips, establishing a shared conditioning format between planning and visual prediction.

% For each segmented video clip, we use GPT-5.6-Terra to annotate its detailed visual context and rewrite the corresponding subtask instruction into a structured visual prompt. Grounded in the clip, the rewritten prompt describes the scene configuration, relevant objects, the progression of the interaction, and the resulting visual changes. These annotations expand concise action instructions into detailed descriptions of how the scene evolves during execution, providing explicit supervision for the reasoner to formulate visual prompts suitable for video generation. The same prompts are paired with their corresponding clips to train the generator, establishing a shared conditioning format between planning and visual prediction.

\paragraph{Agentic data filtering.}
Training data from heterogeneous sources vary in quality and coverage. We combine quality filtering with capability-guided selection to preserve the generator's general capabilities while addressing task-specific weaknesses. An open-source MLLM scores individual clips and complete trajectories based on visual quality, operation completeness, and instruction--video alignment, and low-scoring samples are removed.
We further use GPT-5.6-Terra to analyze videos generated by the base model and identify capability gaps associated with particular scenarios or tasks. We select examples covering similar tasks and scenes from the filtered training pool and combine them with randomly sampled data to retain broad coverage.

\paragraph{Interleaved sample construction.}
To construct long-horizon reasoning examples, we preserve the original task goal and temporal ordering of the demonstrations, interleaving visual observations with subtask instructions, structured visual prompts, and operation durations. This organization trains the reasoner to identify the next subtask from the task goal and interaction history, describe its anticipated visual evolution, and estimate its duration. The aligned prompt--video pairs provide supervision for the generator, while the connections between consecutive clips support video continuation across subtask boundaries. Together, these training examples connect task decomposition, visual prompt formulation, and continuous video prediction throughout a multi-step task.
\subsection{Generation Enhanced Understanding (GEU) Data}

GEU training examples are organized as interleaved conversations: given an observation and a question or task instruction, the assistant first produces a world-modeling instruction, receives the corresponding video as visual context, and then generates a reasoning response ending with an answer or spatial output.

\subsubsection{RynnUnify-Phys}
We construct RynnUnify-Phys from the rendered videos and simulation annotations of PhysInOne~\cite{Zhou2026PhysInOneVP} to connect physical future prediction with reasoning over visual evidence. Each selected sequence is split into an observed prefix and a future continuation, with question--answer pairs derived from object trajectories and event annotations in the continuation. These questions probe how objects move and interact over time, covering displacement and final speed, the timing of contact and settling, and the objects and support relations involved in subsequent interactions. To couple these prediction targets with visual continuation, we organize each SFT example as an interleaved sequence of the observed video and question, a world-modeling instruction, the ground-truth continuation video, and a textual reasoning response ending with the answer. This structure pairs the future visual evidence with the reasoning it supports.

A central consideration in constructing the world-modeling instructions is to direct generation toward the physical interaction relevant to the question while leaving its outcome to be predicted. We therefore use GPT-5.6-Luna~\cite{openai2026gpt56luna} to rewrite the world-modeling instructions using timestamped frames from the observed prefix. The reasoner is instructed to ground each request in the relevant objects, visible motion, and support or contact relations, and to ask for continuation through the interaction without specifying how it resolves. The rewritten instructions undergo format checks and a separate input-conditioned audit for factual grounding, interaction coverage, and future-outcome leakage, with review status retained for unresolved cases. The resulting interleaved examples connect question-conditioned physical continuation with subsequent reasoning over visual evidence. We finally use 10,028 examples for supervised fine-tuning and 1,553 numerical questions for reinforcement learning.

\subsubsection{RynnUnify-Mess}

Physical reasoning in cluttered environments requires anticipating how a local intervention propagates through interactions among objects.
Moving a supporting object, for example, may cause other objects to
slide, rotate, or fall, producing a sequence of dependent changes.
We construct RynnUnify-Mess to support learning these physical
dependencies and predicting how a scene evolves under a specified
intervention.

We first construct cluttered tabletop scenes in RoboTwin~\cite{chen2025robotwin} with interacting and partially occluded objects. Starting from a stable configuration, we apply controlled interventions, such as pulling or removing an object, and simulate the resulting dynamics. Throughout each rollout, we record object poses, contact and support relations, and subsequent events, including support loss, falling, and changes in visibility. These records provide structured supervision for the effects of an intervention on other objects in the scene.

To obtain continuous video supervision, we first use GPT-5.6-Terra~\cite{openai2026gpt56terra} to rewrite the video-generation prompts, explicitly describing the scene configuration, the requested intervention, and the expected physical consequences for other objects. We then condition MiniMax-H3 on these rewritten prompts and the simulated initial and final frames in the FL2V setting to generate the corresponding interaction videos. The rewritten prompts are also retained as world-modeling annotations for subsequent training. Using the recorded simulator states and events, we automatically construct questions and initial reference answers about which objects are affected and how their positions, orientations, or visibility change. This process connects the intervention description, the visual evolution of the scene, and the reasoning required to interpret its consequences.

We combine automatic motion and visibility criteria with GPT-based post-filtering to improve data quality. The former prioritize visually discernible events and suppress negligible peripheral motions and weak visibility changes. The latter assess whether the generated video faithfully depicts the requested intervention with plausible physical interactions, rejecting videos with artifacts such as teleportation, severe interpenetration, or unrelated instability. For retained videos, we additionally check reference answers against the visible outcomes and correct inconsistent descriptions. Each resulting training example combines an initial observation and intervention question with a video-generation prompt, an intermediate video rollout, and an answer, connecting physical prediction with the interpretation of its visual consequences.

\subsubsection{RynnUnify-Spatial}

RynnUnify-Spatial covers position, direction, distance, and size reasoning through category-specific QA construction and a shared video-generation and reasoning-annotation pipeline. We construct the position subset from ScanNet++~\cite{yeshwanth2023scannetpp} using camera intrinsics and extrinsics, object 3D bounding boxes, and per-view visibility measured by visible proportions and pixel counts. We consolidate semantically similar object labels and filter categories unsuitable for position reasoning. Rule-based templates construct four question types: object--camera position, camera--camera position, camera motion, and object--object position. All four types use reference objects visible in both images as common cross-view references and apply question-specific geometric or visibility constraints. Object--camera questions select an object that is absent or only slightly visible in the queried view but clearly visible in the other. Camera--camera questions ask for one camera's position in the other's reference frame. Camera-motion questions constrain camera displacement and isolate a dominant yaw or pitch change to distinguish left, right, upward, and downward rotation. Object--object questions relate objects across the two views using either a specified camera frame or a horizontal compass frame defined by a reference object pair. Answers are computed from scene geometry, with balanced sampling across question types, reference-view configurations, answers, and scenes. The position subset contains 100,000 examples: 99,200 for training and 800 for testing, with 200 test examples per question type. For direction, distance, and size, we sample RynnBrain's video-based spatial perception data~\cite{damo2026rynnbrain} and reformulate them as single- or dual-frame QA examples. Qwen3.8-27B extracts keyframes, assesses visual overlap and semantic relevance between candidate frames, and scores question-answering difficulty. These scores guide filtering and sampling across the three reasoning dimensions.

Across all four categories, initial video-generation prompts are constructed from question-type-specific templates. For position examples, object--camera and camera--camera prompts start from the view defining the reference frame, camera-motion prompts follow the queried motion direction, and object--object prompts randomly use either transition order. Direction examples follow the same view-transition strategy, while size and distance prompts guide the camera to move closer to the target object. Following MiniMax-H3's prompt-expansion guidance, GPT-5.6-Luna~\cite{openai2026gpt56luna} expands the initial prompts using the original images. The expanded prompts and image pairs condition MiniMax-H3~\cite{minimax2026h3} to generate intermediate visual context in FL2V mode, while a portion of the size examples use single-image I2V generation. GPT-5.6-Luna then receives the original visual inputs, question and answer options, video-generation prompt, generated video, and correct answer to produce a chain-of-thought annotation. Each training example links a spatial question, a world-modeling prompt, an intermediate video, and a reasoning response ending with the answer. The resulting data supervise the reasoner's prompt formulation and video-conditioned spatial reasoning, while providing aligned visual inputs, prompts, and videos for generator training in the corresponding mode.

\subsubsection{RynnUnify-ActionReasoning}

We construct RynnUnify-ActionReasoning for embodied planning and predictive world modeling from the physics-aware planning demonstrations in RynnBrain~\cite{damo2026rynnbrain}, including AgiBot~\cite{bu2025agibot_arxiv}, Galaxea~\cite{galaxea_g0}, RoboCasa~\cite{robocasa2024}, RoboCOIN~\cite{wu2025robocoin}, RoboMIND~\cite{wu2025robomind}, RoboTwin~\cite{Mu_2025_CVPR}, SoftFold~\cite{zheng2025x}, and self-captured data. Episodes are divided into adjacent observed-context and future-continuation clips of 2--6 seconds. Inspired by ERQA~\cite{team2025gemini}, the dataset covers seven capabilities: action-primitive selection, translation and reach, action-intent recognition, push direction, rotation and orientation, metric action, and affordance and feasibility. Generation is conditioned on the target category to improve coverage of underrepresented rotation, metric, and affordance questions; push-direction examples are restricted to visually unambiguous pushing transitions.

For each clip pair, GPT-5.6-Luna~\cite{openai2026gpt56luna} uses frames near the observed--continuation boundary to generate a category-specific planning question, a world-modeling instruction, reasoning, and an answer grounded in the visible transition. After selection, world-modeling instructions are refined using the paired frames and question to add motion detail while preserving the target action and excluding unsupported properties, absolute measurements, and hidden states. Each SFT example interleaves the observed video and question, the world-modeling instruction, the ground-truth continuation video, and a reasoning response ending with the answer.
Quality control combines independent multimodal judging for action-primitive, translation, intent, and push questions with applicability scoring for rotation, metric, and affordance questions. The former checks visual grounding, category alignment, answer correctness, reasoning consistency, and instruction--video agreement; the latter assesses whether the transition provides sufficient evidence for the requested reasoning. We retain the top 2{,}000 examples per category, yielding 14{,}000 examples split into 13{,}300 training and 700 test examples (1{,}900 and 100 per category). A balanced 200-example subset supports efficient evaluation.

\subsubsection{RynnUnify-Affordance and Area}
We construct RynnUnify-Affordance and RynnUnify-Area from a subset of RynnBrain-Planning~\cite{damo2026rynnbrain}, which decomposes long-horizon manipulation tasks into temporally ordered atomic subtasks. Each subtask combines a language instruction with spatial annotations, including target object bounding boxes, affordance points, and placement area points. We use its streaming format, in which successive visual observations are interleaved with these spatially grounded subtask instructions.

Using the subtask boundaries and their associated affordance and area annotations, we segment the original videos into individual manipulation clips and retain the corresponding instructions and spatial labels. For RynnUnify-Affordance, we construct task-conditioned questions that ask the model to identify an interaction point on the target object. For RynnUnify-Area, the questions require localizing the target region specified by the subtask, such as the destination for object placement. To support subsequent video-generation training, we further use GPT-5.6-Terra~\cite{openai2026gpt56terra} to produce world-modeling descriptions from the subtask video clips, detailing the scene configuration, the performed operation, and the resulting state changes. These descriptions provide textual conditions for generating the corresponding interaction videos. Each resulting sample associates a visual observation and task instruction with a world-modeling description, an execution clip, and a spatial annotation, linking video prediction with task-conditioned localization. This process yields 20,039 affordance samples and 13,774 area samples, totaling 33,813 training examples.

\subsubsection{RynnUnify-Trajectory}

We construct RynnUnify-Trajectory from RH20T~\cite{fang2024rh20t} manipulation videos to connect task-conditioned trajectory prediction with visual evidence of action execution. Each example pairs the initial frame of a manipulation clip with a task instruction and asks the model to predict the corresponding two-dimensional trajectory in image space. The target trajectory is represented as an ordered sequence of at most ten points, with coordinates normalized to the range of 0--1000. This representation provides spatial supervision for the motion associated with the requested interaction.

To support interleaved reasoning and video generation, we organize each example as an initial observation and instruction, a world-modeling description, the corresponding execution video, and a textual reasoning response ending with the trajectory. The execution clips are at most eight seconds long and sampled at 24 frames per second. We initially use Qwen-VL-Max to generate world-modeling descriptions from the clean initial frame, describing the scene and intended motion. We further rewrite these descriptions using four frames from the execution clip as visual evidence. The rewritten descriptions anchor the scene to the initial observation and specify the progression of the interaction through shot descriptions and an integer-valued target duration. These descriptions provide textual conditions for generating the aligned execution videos, while the videos supply intermediate visual evidence for subsequent trajectory prediction.

We exclude examples whose projected trajectories have a total path length below 20 pixels, including motions that occur primarily in depth and provide little visible displacement in the image plane. We also validate the answer format, coordinate range, and number of trajectory points, and check sample identifiers for uniqueness. The resulting dataset contains 9,892 examples, split into 9,592 training examples and 300 test examples with no overlapping sample identifiers. Each retained example therefore links a task-conditioned visual prediction with an explicit spatial description of the resulting motion.

%% file: sec/4_Training.tex
\section{Training}
\begin{figure*}[t]
    \centering
    \includegraphics[width=\textwidth]{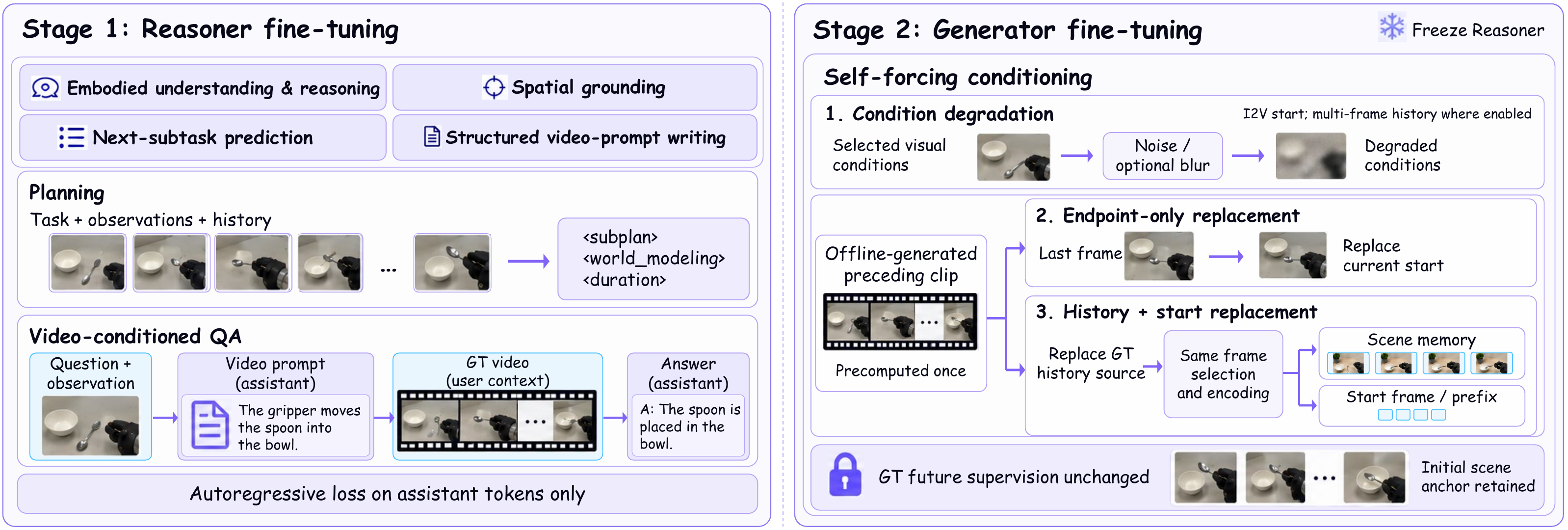}
     \caption{\textbf{Two-stage pretraining of RynnUnify.} Stage 1 fine-tunes the reasoner for embodied understanding, spatial grounding, next-subtask prediction, and structured video-prompt generation. Stage 2 freezes the reasoner and fine-tunes the generator with self-forcing through condition degradation and offline replacement of starting frames and visual history, while retaining ground-truth future videos as supervision.}   
     \label{fig:pretraining}
\end{figure*}

Figure~\ref{fig:pretraining} summarizes the two-stage pretraining procedure: we first fine-tune the reasoner, then freeze it and fine-tune the generator.

\subsection{Stage 1: Reasoner Fine-tuning}
\subsubsection{Training Objectives}
We fine-tune the reasoner inherited from Cosmos3~\cite{cosmos3} to develop four complementary capabilities: spatio-temporal understanding, physical-world grounding, subtask planning, and video-prompt formulation. Following RynnBrain~\cite{damo2026rynnbrain,li2026rynnbrain11capablegeneralizable}, video-based understanding serves two purposes: maintaining broad embodied understanding and reasoning capabilities, and preparing the model to extract and reason over task-relevant information from predicted videos. The latter provides a foundation for subsequent generation-enhanced embodied understanding (GEU), where video predictions serve as intermediate visual evidence for reasoning about spatial relationships, physical interactions, and their consequences before producing an answer. Physical grounding connects these capabilities with explicit objects, interaction regions, and trajectories. We further train the reasoner to determine executable subtasks from the current observation and task history, and to formulate detailed video-generation prompts. Motivated by our experience with MiniMax-H3~\cite{minimax2026h3}, we explicitly supervise prompt formulation to provide the generator with descriptions of both the current scene and its anticipated visual evolution.

\subsubsection{Training Format}
To preserve broad embodied understanding and physical grounding, we follow the multimodal training format of RynnBrain~\cite{damo2026rynnbrain}, an embodied foundation model that jointly learns spatio-temporal understanding, spatial localization, and planning. Images and videos are represented as ordered frame sequences, while assistant responses contain natural language and, when required, spatial coordinates for objects, interaction regions, and trajectories. This shared format supports reasoning over temporal observations while grounding the outputs in explicit elements of the physical scene.

Long-horizon visual planning requires the reasoner to determine the next operation from task progress and provide the generator with a clear description of what should happen during its execution. We therefore jointly supervise subtask prediction and video-prompt formulation. Given the task goal, initial and current observations, and preceding subtask history, the reasoner predicts an executable subplan together with a structured visual prompt describing the relevant scene context, object motions, and anticipated state changes. The subplan specifies the intended operation, while the visual prompt translates that intention into detailed conditions for video generation. We additionally supervise the operation duration to specify the temporal extent of the prediction:
\begin{lstlisting}[style=rynnformat]
USER [context; no loss]:
[HIGH-LEVEL TASK] Task instruction.
[INITIAL STATE] Initial observation.
[PREVIOUS STEPS] Subtask history from the three preceding steps.
[CURRENT OBSERVATION] Current observation.

ASSISTANT [supervised]:
<subplan>Next executable subtask.</subplan>
<world_modeling>
Detailed scene context and expected visual evolution during the operation.
</world_modeling>
<duration>Estimated duration in seconds.</duration>
\end{lstlisting}

Generation-enhanced embodied understanding requires the reasoner to formulate a task-relevant visual prediction and interpret the temporal evidence it provides. To supervise both abilities, we organize these examples into conversations that connect prompt formulation with video-conditioned reasoning. Given an observation and a question, the assistant first writes a video-generation prompt specifying the scene and the visual evolution to be predicted. The corresponding ground-truth video is then supplied as a user turn, followed by the assistant's reasoning and answer. The first assistant turn teaches the model to express the visual prediction needed for the question, while the second teaches it to extract and reason over evidence in the provided video. This supervision prepares the reasoner to use video predictions as intermediate context during subsequent generation-enhanced embodied understanding:
\begin{lstlisting}[style=rynnformat]
USER [context; no loss]:
[VISUAL INPUT] Images or video frames.
[QUESTION] Embodied question.

ASSISTANT [supervised]:
<world_modeling>
Scene context and the requested visual prediction.
</world_modeling>
<duration>Estimated duration in seconds.</duration>

USER [context; no loss]:
[VIDEO] Corresponding ground-truth video frames.

ASSISTANT [supervised]:
<thinking>Reasoning based on the video.</thinking>
<answer>Final answer.</answer>
\end{lstlisting}

\subsubsection{Optimization}
We optimize the reasoner using a standard autoregressive loss over assistant response tokens:
\begin{equation}
\mathcal{L}_{\mathrm{R}}
=
-\sum_{i\in\mathcal{I}_{\mathrm{R}}}
\log p_{\theta_{\mathrm{R}}}(s_i \mid \mathbf{s}_{<i}),
\end{equation}
where $\mathbf{s}$ denotes the multimodal conversation, $\theta_{\mathrm{R}}$ denotes the reasoner parameters, and $\mathcal{I}_{\mathrm{R}}$ indexes the assistant response tokens. Both prompt-generation and answer-generation turns are supervised; system and user turns, including their visual inputs, are excluded from the loss. Only the reasoner is optimized in this stage. The reasoner SFT hyperparameters are summarized in Table~\ref{tab:training_hyperparameters}.
\subsection{Stage 2: Generator Fine-tuning}

\label{sec:generator_training}

Following reasoner fine-tuning, we freeze the reasoner and optimize the generator to translate structured visual prompts and scene context into video predictions. Beyond modeling individual operations, this stage aims to maintain scene consistency across successive subtasks and improve robustness to imperfect visual conditions during long-horizon rollout. We jointly train complementary generation tasks with cross-chunk scene memory and self-forcing strategies.

\subsubsection{Training Tasks}
We combine human and robot manipulation videos with videos for embodied perception and reasoning to jointly train image-to-video generation (I2V), first-and-last-frame-to-video generation (FL2V), and cross-chunk video continuation. These tasks learn to predict subsequent changes from an observation, generate transitions between specified visual states, and extend an interaction from preceding visual history, respectively. Each planning clip is paired with its annotated subtask instruction, a detailed visual prompt describing the anticipated scene evolution, and the operation duration. These aligned text--video pairs connect the structured descriptions learned by the reasoner with the visual predictions expected from the generator.

\subsubsection{Training with Cross-Chunk Scene Memory}
For planning examples, we construct visual conditions from the initial observation and preceding clips within the same episode, following Section~\ref{sec:architecture}. Four independently encoded reference frames, comprising the initial observation and selected historical frames, provide persistent scene information before the main video tokens. Cross-chunk continuation additionally uses a continuous prefix extracted from the preceding clip's tail at the beginning of the main video sequence. These conditions provide both sparse references to earlier scene appearance and layout and continuous context for recent motion and interactions.

During training, we sample from admissible prefix lengths of $k\in\{2,4,8\}$ latent frames. Under the causal VAE's temporal compression scheme, these correspond to the last $1+4(k-1)$ video frames of the preceding clip, yielding prefixes of 5, 13, or 29 frames. We concatenate the selected prefix with the current clip, including the shared boundary time step only once. The combined main video contains at most 241 frames, and the prefix occupies at most half of its latent frames to retain sufficient prediction supervision. The textual duration describes only the current operation and excludes the historical prefix. At inference, we use the final 13 frames of the preceding generated clip as the prefix; if the clip is too short, we perform I2V generation from its final frame while retaining the scene-memory conditions.

\subsubsection{Self-forcing Training}
During long-horizon rollout, the starting observation and recent visual history of each subsequent clip depend on preceding predictions. Training exclusively with ground-truth conditions therefore leaves the generator unprepared for visual errors that accumulate across successive predictions. Inspired by Self Forcing~\cite{huang2025selfforcing}, we employ three complementary strategies to vary the visual conditions encountered during training while retaining ground-truth supervision for the continuation.

\paragraph{Condition degradation.} For selected I2V examples, we perturb the starting condition frame in pixel space before VAE encoding. Specifically, we add Gaussian noise with a randomly sampled strength and optionally apply downsampling--upsampling blur. These perturbations expose the generator to imperfect starting observations and encourage robust prediction when the conditioning image contains noise or loses visual detail.

\paragraph{Offline endpoint replacement.} We generate preceding video clips offline using a fixed generator checkpoint and cache their predictions. The cache is prepared once and reused throughout training without refreshing it as the generator parameters change. For examples with an available cached prediction, we replace the current clip's ground-truth starting observation with the final frame of the corresponding generated clip. This trains the generator to continue from its own predicted scene states. Examples without an available cached prediction retain their original observations.

\paragraph{Joint history and endpoint replacement.} To further expose the generator to self-generated visual history, we extend replacement from the starting frame to the preceding videos used for condition construction. For examples using this strategy, cached generated clips replace the corresponding ground-truth historical videos, and the final frame of the same generated predecessor supplies the current starting observation. Scene-memory references and temporal prefixes are then constructed using the procedures in Section~\ref{sec:architecture}, preserving the existing visual input organization. This ensures that the recent history and its resulting starting observation originate from the same prediction, allowing the model to learn continuation from generated historical context.

Across these strategies, the ground-truth initial observation remains a persistent scene reference, and the current clip's ground-truth continuation remains the supervision target. Scene memory supplies visual information retained across steps, while self-forcing trains the generator to use this information under imperfect conditions.

\subsubsection{Optimization}
We retain the flow-matching objective of the Cosmos3 generator~\cite{cosmos3}, applying supervision only to the main video positions that require prediction:
\begin{equation}
\mathcal{L}_{\mathrm{G}}
=
\mathbb{E}\left[
\frac{1}{|\Omega|}
\sum_{j \in \Omega}
\left\|
v_{\theta_{\mathrm{G}}}
\left(\mathbf{z}_{t},t,\mathbf{c}\right)_{j}
-
\mathbf{v}^{\star}_{t,j}
\right\|_{2}^{2}
\right],
\end{equation}
where $\theta_{\mathrm{G}}$ denotes the generator parameters, $\mathbf{z}_{t}$ the main video latent sequence at flow time $t$, $\mathbf{c}$ the textual and visual conditions, and $\mathbf{v}^{\star}_{t}$ the target flow velocity determined by the ground-truth video latents and sampled noise. The set $\Omega$ indexes the main video positions to be predicted. Fixed conditioning positions and independently encoded scene-memory items are excluded from the loss. We normalize over active prediction positions and average only over visual items containing prediction targets, preventing conditioning inputs from diluting the supervision. Condition degradation and offline replacement modify the conditioning inputs while preserving the ground-truth continuation objective.

For training the generator, we use AdamW with a peak learning rate of $2\times10^{-5}$, $\beta_1=0.9$, $\beta_2=0.95$, $\epsilon=10^{-6}$, and no weight decay. We train for 8,000 optimization steps, linearly increasing the learning rate from zero over the first 800 steps, followed by cosine decay, and clip the global gradient norm to 0.3. For self-forcing training, we apply pixel-space condition degradation to scene-memory-conditioned planning I2V examples with a probability of 0.3. For each selected example, we perturb the frames of the main video clip before VAE encoding, leaving all prediction targets unchanged. With pixel values normalized to $[-1,1]$, we add Gaussian noise with standard deviation $\sigma \sim \mathcal{U}(0,0.15)$. Independently, with a probability of 0.5, we apply downsampling--upsampling blur by sampling a spatial scale $s \sim \mathcal{U}(0.5,1.0)$, downsampling the perturbed frame, and restoring its original resolution using bilinear interpolation. The resulting pixel values are clipped to $[-1,1]$.

\subsection{Stage 3: Agentic Reinforcement Learning}
\label{sec:agentic_rl}
\begin{figure*}[t]
    \centering
    \includegraphics[width=\textwidth]{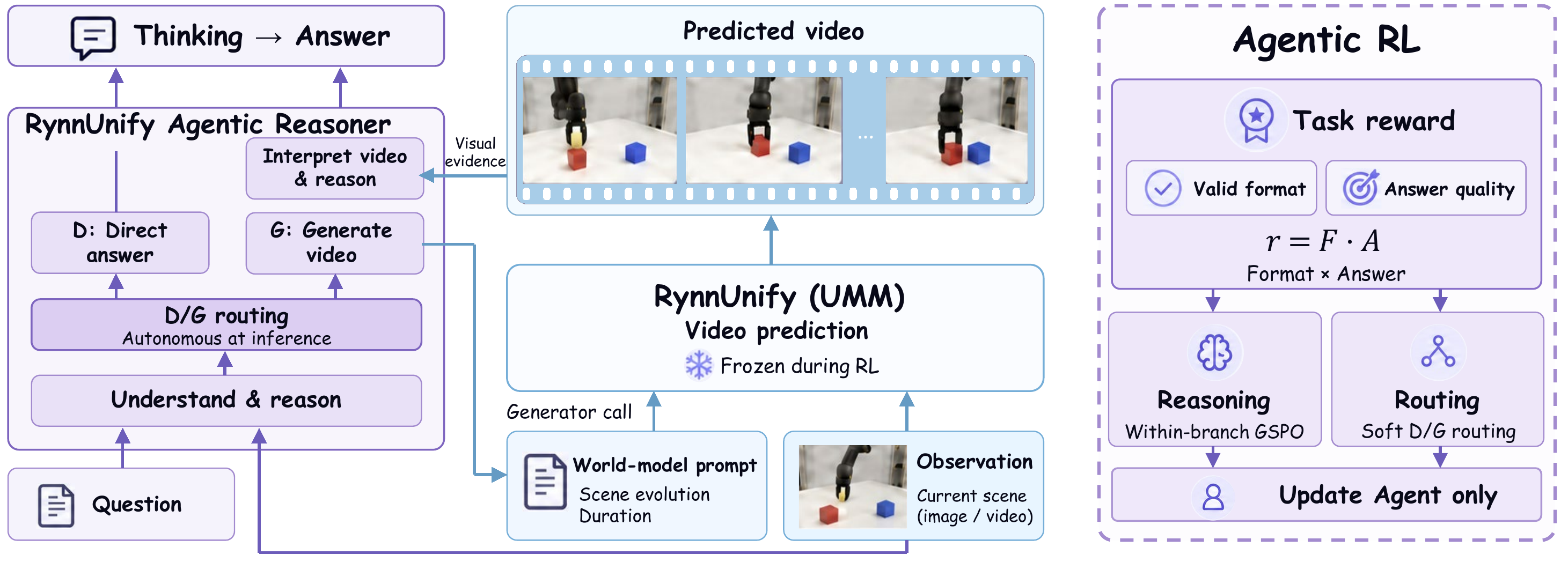}
  \caption{\textbf{Agentic reinforcement learning for generation-enhanced embodied understanding.} The agent dynamically chooses between answering directly and invoking the frozen RynnUnify model to generate visual evidence for further reasoning. Rewards combining format validity and answer quality guide within-branch GSPO and soft D/G routing. Only the agent is updated during reinforcement learning.}
    \label{fig:agentic_rl}
\end{figure*}

% [original]
% To learn when and how visual generation supports reasoning, we propose \textbf{Branch-Conditional Policy Optimization (BCPO)}, an agentic reinforcement learning framework that jointly optimizes reasoning behavior and generator invocation. BCPO combines two complementary learning signals: within-branch reward comparisons improve reasoning along each interaction path, while between-branch reward differences teach the reasoner to select the path that benefits the task. This formulation connects the decision to generate visual evidence with the ability to request and interpret it.

% [revised]
To learn when and how visual generation supports reasoning, we propose \textbf{Branch-Conditional Policy Optimization (BCPO)}, an agentic reinforcement learning framework that jointly optimizes reasoning behavior and generator invocation (Figure~\ref{fig:agentic_rl}). BCPO combines two complementary learning signals: within-branch reward comparisons improve reasoning along each interaction path, while between-branch reward differences teach the reasoner to select the path that benefits the task. This formulation connects the decision to generate visual evidence with the ability to request and interpret it.

Given a question and its visual observations $x$, the reasoner chooses between direct reasoning ($D$) and generation-assisted reasoning ($G$). The direct branch produces a reasoning response and final answer from the original input. The generation-assisted branch follows a request--observe--answer interaction: the reasoner specifies a world-modeling instruction and duration, receives a video from the frozen generator,
and incorporates the returned visual evidence into its response. Task rewards supervise the resulting trajectories, updating the reasoner's generation requests and answers while keeping
the generator fixed. BCPO explores both branches for each routable question, sampling four direct and four generation-assisted trajectories. Within each branch, relative rewards identify effective reasoning behaviors under the same interaction path. Across branches, the difference in mean reward provides a learning target for generator invocation. The reasoner therefore learns both how to reason along each path and how to anticipate the benefit of generation from the original input. At inference time, it autonomously selects a branch and executes the corresponding reasoning process.

\subsubsection{Optimization Objective}
\label{sec:rl_optimization}

\paragraph{Within-branch optimization.}
Based on a sequence-level GSPO~\cite{gspo} objective, BCPO optimizes direct and generation-assisted trajectories in separate groups. For trajectory $i$ in branch $z$, the normalized advantage is
\begin{equation}
    \widehat{A}_{z,i}
    =
    \frac{r_{z,i}-\mu_z}{\sigma_z+\epsilon},
    \label{eq:rl_branch_advantage}
\end{equation}
where $r_{z,i}$ is the trajectory reward, and $\mu_z$ and $\sigma_z$ are the reward mean and standard deviation within that branch. This comparison favors successful reasoning behaviors under the same interaction path. In the generation-assisted branch, the request and subsequent response share the trajectory-level feedback, allowing both to be optimized toward final answer quality.

Let $\rho_{z,i}$ denote the sequence-level importance ratio, computed as the exponential of the mean token log-probability difference between the current and rollout policies. The policy objective is
\begin{equation}
\begin{aligned}
    \mathcal{L}_{\mathrm{GSPO}}
    =-\mathbb{E}_{x,z,i}\Big[
    \min\Big\{
        &\rho_{z,i}\widehat{A}_{z,i},\\
        &\operatorname{clip}
        \big(\rho_{z,i},
        1-\varepsilon_{\mathrm{low}},
        1+\varepsilon_{\mathrm{high}}\big)
        \widehat{A}_{z,i}
    \Big\}\Big].
    \label{eq:rl_gspo}
\end{aligned}
\end{equation}
The objective supervises the reasoner-generated request and response tokens, with generated videos serving as observations. Sequence-level clipping controls policy updates across trajectories of different lengths.

\paragraph{Learning when to generate.}
We train invocation decisions from the mean reward difference between the two paths. For each question, we construct a soft target for generator invocation:
\begin{equation}
    q_G(x)
    =
    \operatorname{sg}\left[
        \operatorname{sigmoid}\left(
            \frac{\mu_G-\mu_D}{T_{\mathrm{route}}}
        \right)
    \right],
    \label{eq:rl_route_target}
\end{equation}
where $\operatorname{sg}$ denotes stop-gradient and $T_{\mathrm{route}}$ controls target smoothness. A higher generation-assisted reward increases the target invocation probability, while similar branch rewards yield a balanced preference.

The reasoner predicts its invocation probability from the two route-token logits under the original input. We fit this prediction to $q_G(x)$ using a masked binary cross-entropy loss $\mathcal{L}_{\mathrm{route}}$, applying supervision to valid branch comparisons with informative reward feedback. This teaches the reasoner to anticipate the benefit of generation from the question and observations available before tool use.

\paragraph{Joint optimization.}
The complete objective combines branch-specific reasoning, invocation learning, and policy regularization:
\begin{equation}
    \mathcal{L}_{\mathrm{RL}}
    =
    \mathcal{L}_{\mathrm{GSPO}}
    +
    \lambda_{\mathrm{route}}\mathcal{L}_{\mathrm{route}}
    +
    \beta\mathcal{L}_{\mathrm{KL}}
    -
    \eta\mathcal{H}.
    \label{eq:rl_total_loss}
\end{equation}
Here, $\mathcal{L}_{\mathrm{KL}}$ regularizes the policy against the frozen initial reasoner, and $\mathcal{H}$ is policy entropy. Within-branch optimization improves how the reasoner acts along each path, while routing supervision determines which path to select. Detailed loss definitions and hyperparameters are provided in Appendix~\ref{app:rl}.

\subsubsection{Task-Grounded Rewards}
\label{sec:rl_rewards}

Both paths share the same task-grounded answer objective, making their reward difference directly relevant to invocation learning. For routable examples, we combine interaction-protocol validity with task-specific answer quality:
\begin{equation}
    r(\tau)
    =
    F(\tau)\,A_k(\widehat{y},y^{\star}),
    \qquad F(\tau)\in\{0,1\},\quad A_k\in[0,1],
    \label{eq:rl_task_reward}
\end{equation}
where $k$ denotes the task, and $\widehat{y}$ and $y^{\star}$ are the predicted and reference answers. For numerical and geometric tasks, a normalized error $e_k$ is mapped to a continuous reward:
\begin{equation}
    A_k=C(e_k),
    \qquad
    C(e)=
    \left[\max\left(0,1-\frac{e}{\delta}\right)\right]^3,
    \qquad \delta=0.30.
    \label{eq:rl_reward_curve}
\end{equation}
Numerical physical and spatial questions use relative error; affordance and area localization use symmetric nearest-point distance; and trajectory prediction uses discrete Fr\'echet distance normalized by reference arc length. The shared shaping function provides graded feedback for numerical precision and geometric alignment. Discrete and textual answers use option matching, normalized answer matching, or task-specific semantic judgments. Mess uses a level-based assessment of predicted interaction outcomes. The complete reward definitions are given in Appendix~\ref{app:rl_rewards}.

\subsubsection{Training Data and Recipe}
\label{sec:rl_recipe}

% [original]
% We train on 22,007 questions spanning embodied perception, physical prediction, spatial reasoning, interaction outcomes, and localization, with 883 questions for in-training validation. Each update uses 64 questions and eight rollouts per question, yielding 512 trajectories. Routable questions use four direct and four generation-assisted rollouts, while native multi-image examples use eight direct rollouts. Training responses are sampled at temperature 1.0 with a limit of 2,048 generated tokens per reasoner turn. We initialize from the preceding trained model and optimize the reasoner for one epoch using Adam at a constant learning rate of $10^{-6}$, keeping the generator and reference policy fixed. We use Qwen3.8-27B-FP8~\cite{qwen38} for task-specific semantic judgments, alongside deterministic scorers for structured answers. Validation is performed every ten updates with greedy decoding and autonomous D/G selection. Full training settings and interaction protocols are provided in Appendix~\ref{app:rl_recipe}.

% [revised]
We train on 22,007 questions spanning embodied perception, physical prediction, spatial reasoning, interaction outcomes, and localization, with 883 questions for in-training validation. Each update uses 64 questions and eight rollouts per question, yielding 512 trajectories. Routable questions use four direct and four generation-assisted rollouts, while native multi-image examples use eight direct rollouts. Training responses are sampled at temperature 1.0 with a limit of 2,048 generated tokens per reasoner turn. We initialize the agent from RynnUnify's fine-tuned 8B reasoner and optimize the agent for one epoch using Adam at a constant learning rate of $10^{-6}$, keeping the generator and reference policy fixed. We use Qwen3.8-27B-FP8~\cite{qwen38} for task-specific semantic judgments, alongside deterministic scorers for structured answers. Validation is performed every ten updates with greedy decoding and autonomous D/G selection. Full training settings and interaction protocols are provided in Appendix~\ref{app:rl_recipe}.

%% file: sec/5_Specific_Task_Supervised_Finetuning.tex
\section{Post-training for the World-Action Model (WAM)}
\label{sec:wam_post_training}

We extend RynnUnify to a world-action model (WAM) for robot control. We train the model in three stages.

\subsection{Model Architecture}
\label{sec:wam_architecture}

\begin{figure*}[t]
    \centering
    \includegraphics[width=\textwidth]{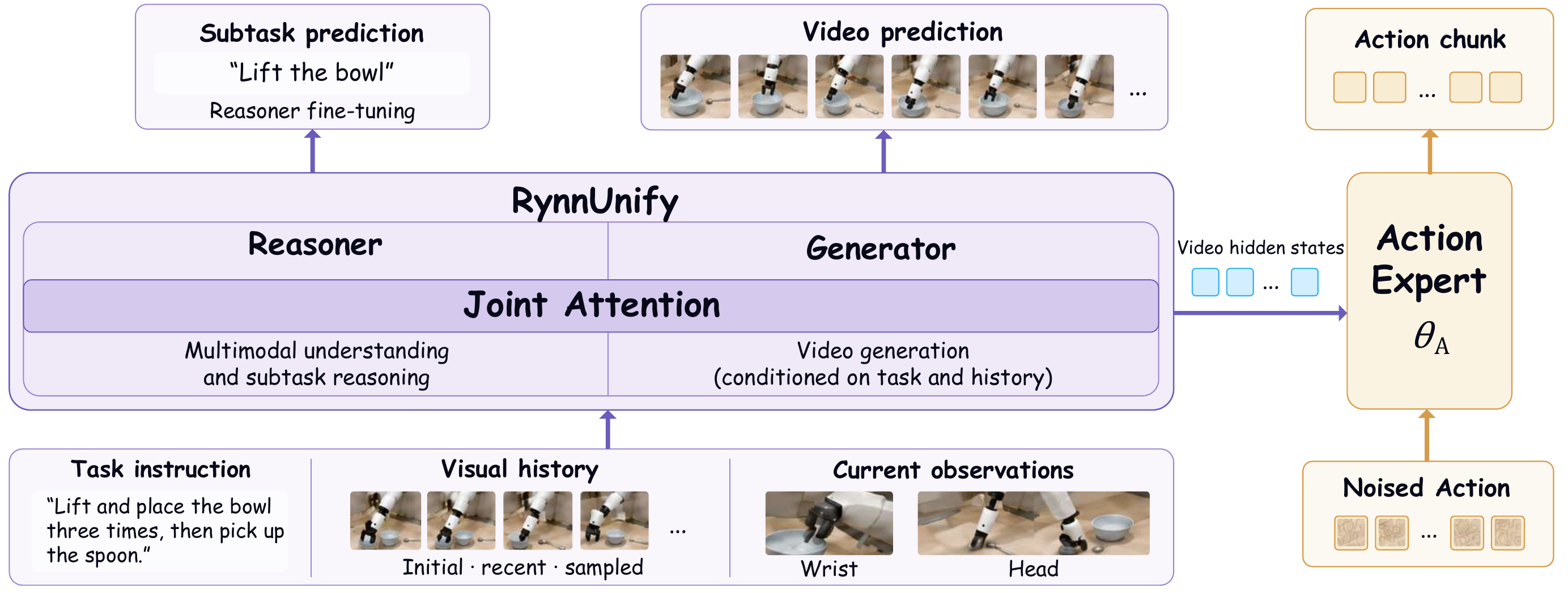}
    \caption{\textbf{Extending RynnUnify to a World-Action Model.}
    The reasoner uses the task instruction, current observations, and
    visual memory to produce a subtask latent. This latent conditions
    the generator through layerwise attention. An action head reads
    the generator's video features to predict action chunks. An
    auxiliary camera-pose head provides geometric supervision during
    training.}
    \label{fig:agent}
\end{figure*}

Figure~\ref{fig:agent} illustrates the WAM architecture. Building on RynnUnify, the reasoner takes the task instruction, current RGB observations, and visual history as input and produces a continuous subtask latent. The generator uses this latent to predict future video, while an action head predicts robot actions from the generator's video features. 

\paragraph{Visual memory.}
At step $t$, the reasoner receives the instruction $\ell$, current observations $\mathcal{O}_t$, and visual history $\mathcal{H}_t$. The current observations contain left, wrist, and right camera views. For RoboCasa365~\cite{robocasa365}, the history includes one left-view image from the start of the episode, up to five earlier left-view images, and up to three
earlier wrist images sampled at different time intervals. Together with the current views, the reasoner uses at most 12 images.

\paragraph{Subtask-conditioned video prediction.}
The reasoner produces a continuous subtask latent $\mathbf{h}^{\mathrm{plan}}_t$ from the instruction and visual context. This latent conditions the generator through layerwise attention. Following RynnUnify, reasoner tokens use causal attention, while generation tokens attend to both reasoner and generation tokens. The current camera views serve as a fixed visual condition, and the generator predicts the future video latents.

\paragraph{Video features and action head.}
We perform a single video denoising step, then use another generator forward pass to collect video hidden states from all 36 layers. An 18-block action head reads these features through cross-attention. Each block attends to two adjacent generator layers, with action tokens as queries and video features as keys and values. The head uses flow matching to predict the velocity field for an action chunk $\mathbf{A}_t\in\mathbb{R}^{48\times12}$.

\subsection{Training}
\label{sec:wam_training}

\paragraph{Training data.} We initialize the model from pretrained RynnUnify. Robot adaptation uses approximately 1,600 hours of atomic-task trajectories and 177 hours of composite-task trajectories from RoboCasa365~\cite{robocasa365}. The data include task instructions, three synchronized RGB views, robot actions, and subtask and skill annotations. We train Stages I and II on atomic tasks and then repeat them on composite tasks, followed by Stage III on composite tasks. For real-robot adaptation, we initialize from the atomic-task model and train the generator and action head on real-robot demonstrations while keeping the reasoner frozen.

Training samples are constructed according to each stage. Stage I uses task instructions, current observations, visual history, and subtask annotations. For Stage II, each sample contains the current multi-view frame, 48 future frames, and 48 aligned actions at 20 Hz, covering 2.4 seconds. Stage III uses the current observations, future actions, and relative camera poses over the same horizon. 

\paragraph{Stage I: Reasoner training.}
We train the reasoner's language layers using subtask and skill annotations, with the visual encoder and generator frozen. We use token-level cross-entropy on the annotated response. This stage trains the reasoner to identify the current subtask from the instruction, current observations, and visual history.

\paragraph{Stage II: Generator training.}
We freeze the reasoner and train the generator using its subtask latent as a condition. Video and native action tokens interact within the shared  generator. We retain the video and action flow-matching losses of the base
model, with equal weights. The video loss follows Section~\ref{sec:generator_training}.

\paragraph{Stage III: Action-head training.}
We jointly train the generator and the new action head while keeping the reasoner and the generator's original action output layer frozen. The action head uses flow matching conditioned on the video features. Its loss also updates the generator through the video denoising step and feature readout.

% We add a camera-pose head that predicts future camera poses relative to the current views. It reads the same video features and is trained along with the action head. The objective is
% \begin{equation}
%     \mathcal{L}_{\mathrm{III}}
%     = \mathcal{L}_{\mathrm{act}}
%     + 0.01\,\mathcal{L}_{\mathrm{pose}}
%     + 0.005\,\mathcal{L}_{\mathrm{rel}},
%     \label{eq:wam_stage3_loss}
% \end{equation}
% where $\mathcal{L}_{\mathrm{act}}$ is the action flow-matching loss, $\mathcal{L}_{\mathrm{pose}}$ measures translation and rotation errors, and $\mathcal{L}_{\mathrm{rel}}$ measures consistency between camera pairs. The SFT hyperparameters for the three training stages are summarized in Table~\ref{tab:training_hyperparameters}. 

We also train an auxiliary camera-pose head to predict future camera poses relative to the current observation. The Stage III objective is
\begin{equation}
    \mathcal L_{\mathrm{III}}
    =\mathcal L_{\mathrm{act}}
     +0.01\,\mathcal L_{\mathrm{pose}}
     +0.005\,\mathcal L_{\mathrm{rel}}.
    \label{eq:vla_stage3_loss}
\end{equation}

% \paragraph{Loss functions.}
For an action chunk $\mathbf A$, we sample Gaussian noise $\boldsymbol\epsilon^{\mathrm a}$ and form $\mathbf A_\sigma=(1-\sigma)\mathbf A+ \sigma\boldsymbol\epsilon^{\mathrm a}$. The action head predicts its flow velocity from $\mathbf A_\sigma$ and the video features $\mathbf H$:
\begin{equation}
    \mathcal L_{\mathrm{act}}=
    \mathbb E_{\sigma,\boldsymbol\epsilon^{\mathrm a}}
    \!\left[\operatorname{MSE}\!\left(
      \mathcal E_\psi(\mathbf A_\sigma,\sigma;\mathbf H),
      \boldsymbol\epsilon^{\mathrm a}-\mathbf A
    \right)\right].
    \label{eq:wam_action_loss}
\end{equation}

For camera $c$, let $\mathbf T_{t,k,c}=\mathbf C_{t,c}^{-1}\mathbf C_{t+k,c}$ denote its pose at future step $k$ relative to the current frame. We measure pose error using
\begin{equation}
    D(\widehat{\mathbf T},\mathbf T)
    =\frac{\|\widehat{\mathbf p}-\mathbf p\|_2^2}{3s_{\mathrm T}^2}
     +\frac{\|\widehat{\mathbf R}-\mathbf R\|_F^2}{9},
    \label{eq:wam_pose_error}
\end{equation}
where $s_{\mathrm T}=0.25\,\mathrm m$. The pose and cross-camera losses are
\begin{equation}
\begin{aligned}
    \mathcal L_{\mathrm{pose}}
      &=\mathbb E_{k,c}\,D(\widehat{\mathbf T}_{t,k,c},\mathbf T_{t,k,c}),\\
    \mathcal L_{\mathrm{rel}}
      &=\mathbb E_{k,c<d}\,D(\widehat{\mathbf E}_{t,k,cd},\mathbf E_{t,k,cd}).
\end{aligned}
    \label{eq:wam_pose_relation_losses}
\end{equation}
Here, $\widehat{\mathbf E}_{t,k,cd}$ and $\mathbf E_{t,k,cd}$ are the relative transforms between camera pair $(c,d)$, formed from the predicted and ground-truth camera poses, respectively, using their known relation at the current frame. All losses are averaged over valid targets; the action loss uses two independent noise--time samples per training example.

\paragraph{Inference.}
At each decision step, we extract the video features and reuse them during action sampling. The action head takes 16 Euler steps to produce a 48-step action chunk, of which the first 16 actions are returned for execution. The camera-pose head is used only in training.

%% file: sec/6_Evaluation.tex
\section{Rynn-VP and Rynn-GEU Benchmarks}
\begin{figure*}[t]
    \centering
    \includegraphics[width=\textwidth]{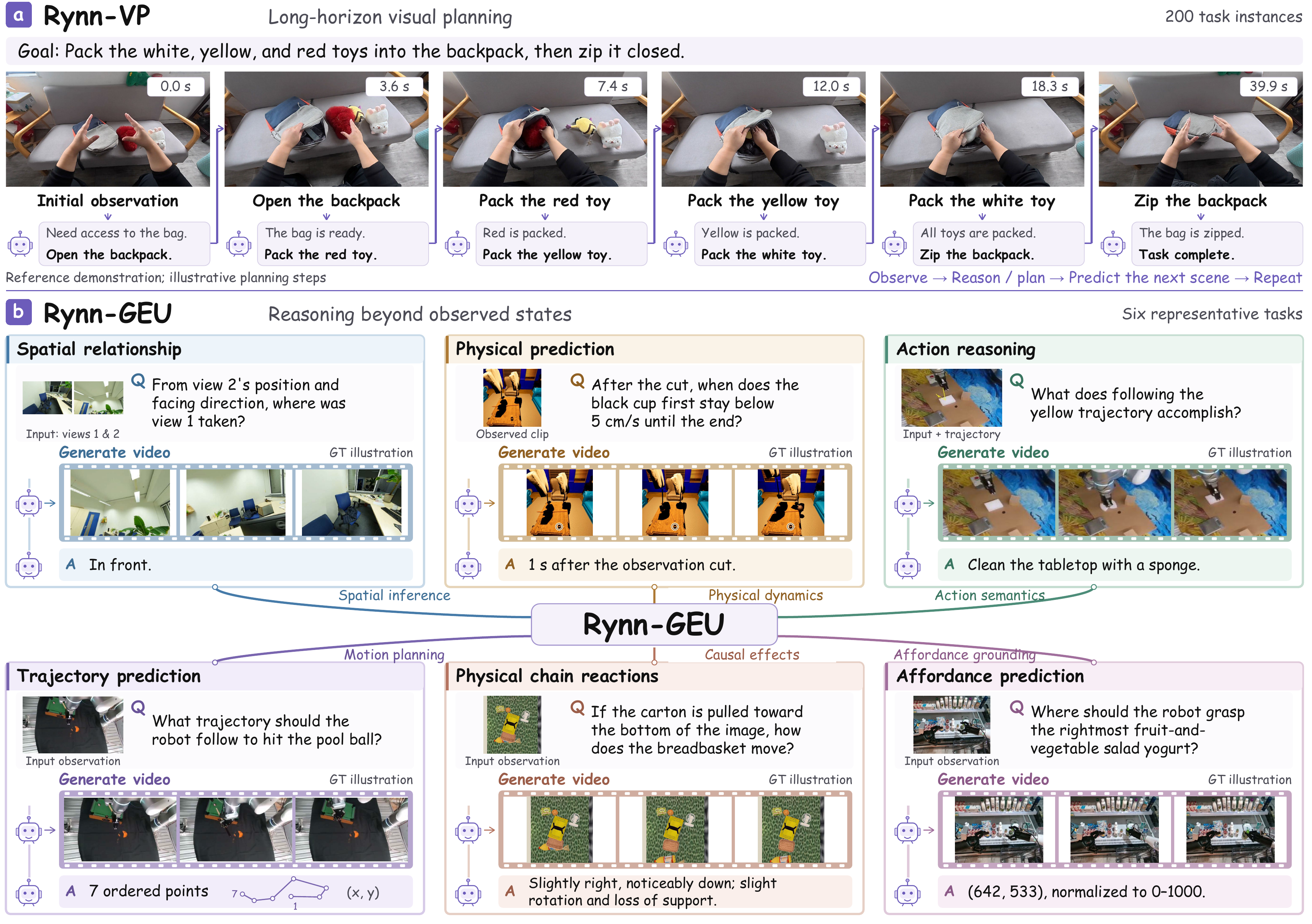}
 \caption{\textbf{Overview of Rynn-VP and Rynn-GEU.} (a) Rynn-VP evaluates long-horizon visual planning across 200 task instances involving interdependent operations. (b) Rynn-GEU evaluates reasoning beyond observed states under both direct and generation-assisted settings, with six representative tasks illustrated here. The videos are ground-truth references, and the planning text is illustrative. }
    \label{fig:bench}
\end{figure*}
\label{sec:rynn_vp}

% [original]
% Reciprocal interaction between visual generation and understanding can help embodied models anticipate task-relevant scene evolution and use predicted outcomes to guide reasoning and planning. Despite extensive exploration of this interaction in general-purpose unified multimodal models~\cite{realunify, uni-mmmu}, dedicated benchmarks in embodied settings remain limited; we therefore introduce Rynn-VP and Rynn-GEU.

% [revised]
Reciprocal interaction between visual generation and understanding can help embodied models anticipate task-relevant scene evolution and use predicted outcomes to guide reasoning and planning. Despite extensive exploration of this interaction in general-purpose unified multimodal models~\cite{realunify,uni-mmmu}, dedicated benchmarks in embodied settings remain limited; we therefore introduce Rynn-VP and Rynn-GEU, summarized in Figure~\ref{fig:bench}.

% Preamble

\subsection{Rynn-VP: Long-Horizon Visual Planning}
\label{sec:rynn_vp_benchmark}

\providecommand{\rynncell}[1]{#1}

\definecolor{rynn}{RGB}{108,92,186}
\renewcommand{\rynncell}[1]{{\small\textcolor{black}{#1}}}

\providecommand{\geucell}[1]{#1}
\renewcommand{\geucell}[1]{{\small\textcolor{black}{#1}}}

Efficiently evaluating embodied planning remains challenging: a high-level task requires interdependent operations that transform an initial scene into a goal-satisfying state, with each operation changing the conditions for subsequent decisions. Next-step prediction measures local decisions but does not establish full-trajectory validity~\cite{egoplanbench,robobrain}. Evaluating complete plans in simulators or with human controllers~\cite{damo2026rynnbrain,Robix} requires environment setup and repeated execution, which can constrain evaluation scale and efficiency.
We view visual planning as a native capability of embodied unified models that interleaves task-advancing subtask reasoning with coherent video prediction, using predicted scene changes to inform subsequent planning.

To evaluate this capability, we introduce Rynn-VP, comprising 200 manually collected egocentric task scenarios. Given an initial image and task instruction, a model generates a video depicting task execution or an ordered sequence of step images. Our \textbf{simulator-free} protocol assesses task progress, operation dependencies, scene continuity, and physical plausibility from these generated rollouts. Reference demonstrations of successful execution and timestamped subtask annotations are reserved for evaluation, specifying task goals and required process evidence while allowing alternative valid execution sequences.

% [original]
% We formalize the scoring criteria and verification procedures as a reusable evaluation Skill, executed by a vision-language model within the Codex harness. The evaluator first examines the reference demonstration and annotations to identify task goals, relevant entities, and dependencies between subtasks. It then inspects the generated visual sequence to assess task progress, intermediate actions, and scene-state transitions. A dedicated hallucination audit further examines object lifecycles, physical interactions, and scene consistency, identifying unsupported object appearance or disappearance, contradictory state changes, physically implausible interactions, and unexplained changes in scene layout. The audit distinguishes clear visual inconsistencies from occlusion or insufficient evidence, and its findings inform the calibration of the corresponding planning and visual realization scores. The resulting structured judgments undergo deterministic score aggregation. We further validate the accuracy of the evaluation procedure through manual review of evidence interpretation, hallucination judgments, and scoring decisions.

% [revised]
We formalize the scoring criteria and verification procedures as a reusable evaluation Skill, executed by GPT-6-sol within the Codex harness. The evaluator first examines the reference demonstration and annotations to identify task goals, relevant entities, and dependencies between subtasks. It then inspects the generated visual sequence to assess task progress, intermediate actions, and scene-state transitions. A dedicated hallucination audit further examines object lifecycles, physical interactions, and scene consistency, identifying unsupported object appearance or disappearance, contradictory state changes, physically implausible interactions, and unexplained changes in scene layout. The audit distinguishes clear visual inconsistencies from occlusion or insufficient evidence, and its findings inform the calibration of the corresponding planning and visual realization scores. The resulting structured judgments undergo deterministic score aggregation. Manual review of the evaluation procedure is described in Appendix~\ref{app:vp_evaluation}.

Rynn-VP measures performance along three dimensions:
\begin{itemize}
    \item \textbf{Planning Intelligence (PI):} Evaluates executable goal achievement, step validity, dependency satisfaction, and plan completeness based on the generated visual evidence.
    \item \textbf{Visual Realization (VR):} Evaluates visible goal completion and action evidence, together with entity consistency, state continuity, consistency across successive video segments, and physical feasibility and plausibility.
    \item \textbf{Visual Quality (VQ):} Characterizes perceptual quality and the clarity of task-relevant visual evidence.
\end{itemize}

All scores are reported on a scale of 0--100. The overall benchmark score combines the calibrated PI and VR scores with VQ:
\begin{equation}
    \mathrm{Score}_{\mathrm{Rynn\text{-}VP}}
    = 0.5\,\mathrm{PI} + 0.4\,\mathrm{VR} + 0.1\,\mathrm{VQ} .
\end{equation}
The evaluation recognizes valid partial progress, and results are reported with the output modality identified to distinguish videos from ordered step images. Detailed scoring rubrics, submetric weights, evidence-sampling procedures, and calibration rules are provided in Appendix~\ref{app:vp_evaluation}.

\subsection{Rynn-GEU: Embodied Understanding with Generated Visual Context}
\label{sec:rynn_geu}

Embodied understanding requires anticipating how actions, physical interactions, and viewpoint changes affect states that are not directly observable in the input. We introduce Rynn-GEU to evaluate this anticipatory reasoning and investigate whether explicit video prediction improves it. The benchmark comprises ten subsets covering spatial relationships (position, direction, distance, and size), physical prediction, chain reactions in cluttered scenes, action reasoning, affordance grounding, area grounding, and trajectory prediction.

\paragraph{Data construction.}
We follow the construction pipelines of the corresponding RynnUnify-GEU training data, with explicit separation between training and evaluation scenes to limit scene memorization. Mess uses RoboTwin2.0~\cite{chen2025robotwin} scenes absent from training, while Spatial, Trajectory, and Action Reasoning exclude scenes that occur in or closely resemble the training data. Phys is constructed from PhysInOne~\cite{Zhou2026PhysInOneVP} using distinct assets and scenes for training and testing. Affordance and Area are constructed from AgiBot-Beta~\cite{bu2025agibot_arxiv}, providing a different source from the RynnBrain-Planning data~\cite{damo2026rynnbrain} used for training.

\paragraph{Future-oriented sample selection.}
To emphasize questions that benefit from reasoning about subsequent states, we further filter the constructed examples using GPT-5.6-Luna~\cite{openai2026gpt56luna}. For each candidate, we compare its answers under the same initial observations and question, with and without access to the reference future video. We primarily retain examples answered well with future visual evidence but poorly without it, while also including a smaller portion answered correctly without the reference video. This process emphasizes questions sensitive to future visual evidence while retaining variation in direct-reasoning difficulty. Because model responses are stochastic, the comparison serves as a selection heuristic rather than a definitive label of whether explicit generation is necessary.

\paragraph{Evaluation metrics.}
We use task-specific metrics on a common 0--100 scale. Action Reasoning uses choice accuracy, while categorical Phys questions use normalized exact matching. For Spatial Position, Direction, and Size, GPT-5.6-Luna compares predictions with reference answers, assigning 100 only when the answer is both correct and complete, and 0 otherwise.

For Spatial Distance, we convert predictions to meters and compute
\begin{equation}
    s_{\mathrm{dist}}
    =
    100\max\left(
        0,\,
        1-\frac{|\hat{d}-d|}{d}
    \right),
\end{equation}
where $\hat{d}$ and $d>0$ are the predicted and reference distances. For numerical Phys questions, we validate the predicted value and its unit compatibility, then compute
\begin{equation}
    s_{\mathrm{phys}}
    =
    100\cdot
    2^{-\left(
        \frac{|\hat{y}-y|}{0.1|y|}
    \right)^2},
\end{equation}
where $\hat{y}$ and $y\neq 0$ are the predicted and reference physical quantities. For either metric, a zero-valued reference receives 100 only when the prediction is also zero.

Affordance and Area evaluate the spatial agreement between predicted and reference point sets:
\begin{equation}
    s_{\mathrm{point}}
    =
    100\exp\left(-5d_{\mathrm{C}}(P,Q)\right),
\end{equation}
where $P$ and $Q$ are the predicted and reference point sets in normalized image coordinates, and $d_{\mathrm{C}}$ is their symmetric mean nearest-neighbor distance. Trajectory evaluates both path geometry and point ordering:
\begin{equation}
    s_{\mathrm{traj}}
    =
    100\exp\left(-d_{\mathrm{F}}(\hat{\mathcal{T}},\mathcal{T})\right),
\end{equation}
where $\hat{\mathcal{T}}$ and $\mathcal{T}$ are the predicted and reference trajectories after coordinate normalization and arc-length resampling, and $d_{\mathrm{F}}$ denotes their discrete Fr\'echet distance.

% [original]
% For Mess, GPT-5.6-Luna assigns 0, 25, 50, 75, or 100 points according to the correctness and coverage of predicted effects on non-manipulated objects, including displacement, rotation, and support or contact changes. Each subset reports its mean per-example score. Detailed geometric distance definitions, parsing rules, and judging prompts are provided in the appendix.

% [revised]
For Mess, GPT-5.6-Luna assigns 0, 25, 50, 75, or 100 points according to the correctness and coverage of predicted effects on non-manipulated objects, including displacement, rotation, and support or contact changes. Each subset reports its mean per-example score.

\section{Experiments}

\begin{table*}[t]
    \centering
    \caption{
        Comparison on public embodied understanding benchmarks without generated-video rollouts.
        \textbf{Bold} and \underline{underlining} mark the best results among UMMs and all compared models, respectively.
        Avg. is the unweighted mean of ten metrics normalized to a 0–100 scale.
    }
    \label{tab:embodied_understanding}

    \footnotesize
    \setlength{\tabcolsep}{1.75pt}
    \renewcommand{\arraystretch}{1.15}

    \begin{tabularx}{\textwidth}{@{}l
        *{6}{>{\hsize=0.8875\hsize\linewidth=\hsize\centering\arraybackslash}X}
        *{3}{>{\hsize=1.3\hsize\linewidth=\hsize\centering\arraybackslash}X}
        *{2}{>{\hsize=0.8875\hsize\linewidth=\hsize\centering\arraybackslash}X}@{}}
        \toprule
        \multirow{2}{*}[-1ex]{\textbf{Model}}
        & \multicolumn{6}{c}{\textbf{Embodied Cognition}}
        & \multicolumn{4}{c}{\textbf{Embodied Localization}}
        & \multirow{2}{*}[-1ex]{\textbf{Avg.}} \\
        \cmidrule(lr){2-7}
        \cmidrule(lr){8-11}
        & \makebox[\linewidth][c]{\scriptsize \makecell[c]{VSI-\\Bench}}
        & \makebox[\linewidth][c]{\scriptsize \makecell[c]{MMSI-\\Bench}}
        & \makebox[\linewidth][c]{\scriptsize \makecell[c]{Mind\\Cube}}
        & \makebox[\linewidth][c]{\scriptsize \makecell[c]{ERQA}}
        & \makebox[\linewidth][c]{\scriptsize \makecell[c]{Robo\\Spatial}}
        & \makebox[\linewidth][c]{\scriptsize \makecell[c]{Emb\\Spatial}}
        & \makebox[\linewidth][c]{\scriptsize \makecell[c]{RefSpatial-\\Bench}}
        & \makebox[\linewidth][c]{\scriptsize \makecell[c]{ShareRobot\\Affordance}}
        & \makebox[\linewidth][c]{\scriptsize \makecell[c]{ShareRobot\\Trajectory$\downarrow$}}
        & \makebox[\linewidth][c]{\scriptsize \makecell[c]{Where2\\Place}}
        & \\[1pt]
        \midrule

        % Open-source understanding models
        \rowcolor{black!5}
        \multicolumn{12}{@{}l@{}}{\rule{0pt}{2.3ex}\textbf{Open-source Understanding Models}} \\

        Qwen3-VL-8B-Instruct~\cite{qwen3vl}\rule[-4pt]{0pt}{13pt}
        & 60.3
        & 29.8
        & 36.8
        & 44.8
        & 58.2
        & 78.5
        & 53.4
        & 37.9
        & 0.37
        & 64.7
        & 52.7 \\

        Qwen3.5-9B~\cite{qwen3.5}\rule[-4pt]{0pt}{13pt}
        & 61.0
        & 13.4
        & 33.1
        & 41.5
        & 54.1
        & 76.9
        & 37.3
        & 7.1
        & 0.53
        & 48.8
        & 42.0 \\

        RynnBrain-8B~\cite{damo2026rynnbrain}\rule[-4pt]{0pt}{13pt}
        & \underline{71.0}
        & \underline{39.6}
        & 56.6
        & 46.8
        & \underline{73.1}
        & 80.4
        & 59.2
        & \underline{44.7}
        & 0.35
        & \underline{75.5}
        & \underline{61.2} \\

        MiMo-Embodied-7B~\cite{MiMo-Embodied}\rule[-4pt]{0pt}{13pt}
        & 48.5
        & 30.2
        & 43.1
        & 46.8
        & 70.3
        & 76.2
        & 48.0
        & 35.6
        & 0.41
        & 63.6
        & 52.1 \\

        Embodied-R1.5-8B~\cite{yuan2026embodiedr15}\rule[-4pt]{0pt}{13pt}
        & 56.1
        & 29.8
        & 27.9
        & 46.0
        & 69.7
        & 78.1
        & 54.2
        & 25.2
        & 0.29
        & 74.0
        & 53.2 \\

        \midrule

        % Closed-source understanding models
        \rowcolor{black!5}
        \multicolumn{12}{@{}l@{}}{\rule{0pt}{2.3ex}\textbf{Closed-source Understanding Models}} \\

        GPT-5.6-Terra~\cite{openai2026gpt56terra}\rule[-4pt]{0pt}{13pt}
        & 58.9
        & 36.9
        & \underline{62.6}
        & \underline{56.5}
        & 57.2
        & 81.1
        & 31.4
        & 31.3
        & 0.29
        & 39.3
        & 52.6 \\

        Claude-Sonnet-5~\cite{anthropic2026sonnet5}\rule[-4pt]{0pt}{13pt}
        & 37.3
        & 28.1
        & 49.6
        & 46.2
        & 41.3
        & 74.3
        & 15.1
        & 2.9
        & 0.47
        & 3.3
        & 35.1 \\

        \midrule

        % Unified multimodal models
        \rowcolor{black!5}
        \multicolumn{12}{@{}l@{}}{\rule{0pt}{2.3ex}\textbf{Unified Multimodal Models}} \\

        BAGEL~\cite{deng2025bagel}\rule[-4pt]{0pt}{13pt}
        & 34.5
        & 31.0
        & 34.7
        & 39.5
        & 41.0
        & 67.0
        & 8.6
        & 3.2
        & 0.88
        & 8.8
        & 28.0 \\

        RxBrain~\cite{liang2026rxbrain}\rule[-4pt]{0pt}{13pt}
        & 43.5
        & 35.8
        & \textbf{47.5}
        & 42.8
        & 60.5
        & 82.3
        & 34.1
        & 11.7
        & 0.49
        & 39.8
        & 44.9 \\

        Cosmos 3 Nano~\cite{cosmos3}\rule[-4pt]{0pt}{13pt}
        & 54.9
        & \textbf{36.2}
        & 40.5
        & \textbf{46.0}
        & 62.1
        & 81.4
        & 53.1
        & 28.0
        & 0.34
        & \textbf{64.0}
        & 53.2 \\

        % RynnUnify: black text, larger font,
        % light purple background and purple borders
        \arrayrulecolor{rynn}
        \specialrule{0.9pt}{3pt}{0pt}
        \rowcolor{rynn!8}
        \rynncell{\rule[-0.5ex]{0pt}{2.8ex}\textbf{RynnUnify (Ours)}}\rule[-4pt]{0pt}{13pt}
        & \rynncell{\textbf{67.1}}
        & \rynncell{35.9}
        & \rynncell{44.2}
        & \rynncell{43.8}
        & \rynncell{\textbf{72.8}}
        & \rynncell{\underline{\textbf{83.0}}}
        & \rynncell{\underline{\textbf{60.3}}}
        & \rynncell{\textbf{41.8}}
        & \rynncell{\underline{\textbf{0.27}}}
        & \rynncell{56.9}
        & \rynncell{\textbf{57.9}} \\
        \specialrule{0.9pt}{0pt}{0pt}
        \arrayrulecolor{black}
    \end{tabularx}
\end{table*}

\begin{table*}[t]
    \centering
    \caption{
        Embodied reasoning on public benchmarks and Rynn-GEU.
        Interleaved Thinking uses generated visual context.
        Agentic Thinking augments RynnUnify-16B with an 8B RL-trained reasoner.
        \textbf{Bold}, \underline{underlining}, and $^{\dagger}$ indicate the best results among UMMs and their reasoning variants, direct-answering models, and all methods, respectively.
    }
    \label{tab:geu_main_results}

    \footnotesize
    \setlength{\tabcolsep}{0.75pt}
    \renewcommand{\arraystretch}{1.15}

    \begin{tabularx}{\textwidth}{@{}l
        >{\hsize=1.15\hsize\linewidth=\hsize\centering\arraybackslash}X
        >{\hsize=1.0\hsize\linewidth=\hsize\centering\arraybackslash}X
        >{\hsize=1.2\hsize\linewidth=\hsize\centering\arraybackslash}X
        *{9}{>{\hsize=0.95\hsize\linewidth=\hsize\centering\arraybackslash}X}
        >{\hsize=1.2\hsize\linewidth=\hsize\centering\arraybackslash}X
        *{2}{>{\hsize=0.95\hsize\linewidth=\hsize\centering\arraybackslash}X}@{}}
        \toprule
        \multirow{3}{*}[-2.25ex]{\textbf{Model}}
        & \multicolumn{4}{c}{\textbf{Public Benchmarks}}
        & \multicolumn{11}{c}{\textbf{Rynn-GEU}} \\
        \cmidrule(lr){2-5}
        \cmidrule(lr){6-16}
        & \multirow{2}{*}[-1.25ex]{{\makecell{MMSI-\\Bench}}}
        & \multirow{2}{*}[-1.25ex]{\makecell{Mind\\Cube}}
        & \multirow{2}{*}[-1.25ex]{ERQA}
        & \multirow{2}{*}[-1.25ex]{\textbf{Avg.}}
        & \multicolumn{3}{c}{Perception}
        & \multicolumn{5}{c}{Spatial}
        & \multicolumn{2}{c}{Physical}
        & \multirow{2}{*}[-1.25ex]{\textbf{Avg.}} \\
        \cmidrule(lr){6-8}
        \cmidrule(lr){9-13}
        \cmidrule(lr){14-15}
        & & & &
        & { Traj.}
        & { Aff.}
        & { Area}
        & { Pos.}
        & { Dist.}
        & { Size}
        & { Dir.}
        & {\makecell{Traj.\\Reas.}}
        & { Physics}
        & { Mess}
        & \\
        \midrule

        % Open-source understanding models: direct answering
        \rowcolor{black!5}
        \multicolumn{16}{@{}l@{}}{\rule{0pt}{2.3ex}\textbf{Open-source Understanding Models}} \\

        Qwen3-VL-8B-Instruct~\cite{qwen3vl}\rule[-4pt]{0pt}{13pt}
        & 29.8 & 36.8 & 44.8 & 37.1
        & 67.3 & 62.3 & 31.7
        & 27.5 & 39.6 & 52.5 & 20.0 & 45.5
        & 15.0 & 7.3
        & 36.9 \\

        % Qwen3.5-9B direct: scores before display rounding, in table column order: 13.4, 33.1, 41.5, 29.3, 73.8, 47.49, 17.47, 42.0, 9.17, 57.5, 17.0, 45.0, 7.21, 15.75, 33.2.
        Qwen3.5-9B~\cite{qwen3.5}\rule[-4pt]{0pt}{13pt}
        & 13.4 & 33.1 & 41.5 & 29.3
        & 73.8 & 47.5 & 17.5
        & 42.0 & 9.2 & 57.5 & 17.0 & 45.0
        & 7.2 & 15.8
        & 33.2 \\

        RynnBrain-8B~\cite{damo2026rynnbrain}\rule[-4pt]{0pt}{13pt}
        & \underline{39.6}\rlap{$^{\dagger}$} & 56.6 & 46.8 & 47.7
        & \underline{79.2} & 60.5 & 28.1
        & 8.0 & 43.5
        & \underline{63.0}\rlap{$^{\dagger}$}
        & \underline{34.5}\rlap{$^{\dagger}$}
        & 40.0
        & 23.0 & 9.5
        & 38.9 \\

        MiMo-Embodied-7B~\cite{MiMo-Embodied}\rule[-4pt]{0pt}{13pt}
        & 30.2 & 43.1 & 46.8 & 40.0
        & 64.8 & 16.0 & 15.1
        & 34.0 & 45.0 & 53.5 & 13.0 & 47.5
        & 5.3 & 18.2
        & 31.2 \\

        Embodied-R1.5-8B~\cite{yuan2026embodiedr15}\rule[-4pt]{0pt}{13pt}
        & 29.8 & 27.9 & 46.0 & 34.6
        & 77.9 & 63.5 & 35.6
        & 28.0 & 43.1 & 52.5 & 19.5 & 44.0
        & 21.9 & 5.3
        & 39.1 \\

        \midrule

        % Closed-source understanding models: direct answering
        \rowcolor{black!5}
        \multicolumn{16}{@{}l@{}}{\rule{0pt}{2.3ex}\textbf{Closed-source Understanding Models}} \\

        GPT-5.6-Terra~\cite{openai2026gpt56terra}\rule[-4pt]{0pt}{13pt}
        & 36.9
        & \underline{62.6}\rlap{$^{\dagger}$}
        & \underline{56.5}\rlap{$^{\dagger}$}
        & \underline{52.0}\rlap{$^{\dagger}$}
        & 77.6 & 53.2 & \underline{42.9}
        & \underline{44.0}\rlap{$^{\dagger}$}
        & 46.5 & 60.0 & 24.0 & 56.0
        & 20.0 & 26.3
        & \underline{45.1} \\

        Claude-Sonnet-5~\cite{anthropic2026sonnet5}\rule[-4pt]{0pt}{13pt}
        & 28.1 & 49.6 & 46.2 & 41.3
        & 70.9 & 20.6 & 16.0
        & 31.5 & 47.5 & 58.0 & 22.2
        & \underline{63.0}\rlap{$^{\dagger}$}
        & \underline{24.0} & \underline{27.1}
        & 38.1 \\

        \midrule

        % UMMs: direct answering followed by generated-video reasoning
        \rowcolor{black!5}
        \multicolumn{16}{@{}l@{}}{\rule{0pt}{2.3ex}\textbf{Unified Multimodal Models}} \\

        BAGEL~\cite{deng2025bagel}\rule[-4pt]{0pt}{13pt}
        & 31.0 & 34.7 & 39.5 & 35.1
        & 4.4 & 26.9 & 3.0
        & 11.5 & 15.0 & 41.0 & 7.0 & 8.0
        & 0.0 & 2.0
        & 11.9 \\

        % BAGEL rollout: scores before display rounding, in table column order: 30.70, 39.62, 44.75, 38.4, 1.4, 22.8, 3.5, 18.5, 10.8, 33.5, 5.5, 10.0, 0.5, 1.9, 10.8.
        \quad + Interleaved Thinking\rule[-4pt]{0pt}{13pt}
        & 30.7 & 39.6 & 44.8 & 38.4
        & 1.4 & 22.8 & 3.5
        & 18.5 & 10.8 & 33.5 & 5.5 & 10.0
        & 0.5 & 1.9
        & 10.8 \\

        RxBrain~\cite{liang2026rxbrain}\rule[-4pt]{0pt}{13pt}
        & 35.8 & 47.5 & 42.8 & 42.0
        & 42.8 & 44.9 & 27.0
        & 24.5 & 40.2 & 44.5 & 21.5 & 33.5
        & 13.7 & 21.0
        & 31.4 \\

        % RxBrain rollout: scores before display rounding, in table column order: 31.80, 45.33, 37.00, 38.0, 33.3, 46.0, 27.0, 24.0, 37.9, 42.5, 20.5, 26.0, 9.6, 19.5, 28.6.
        \quad + Interleaved Thinking\rule[-4pt]{0pt}{13pt}
        & 31.8 & 45.3 & 37.0 & 38.0
        & 33.3 & 46.0 & 27.0
        & 24.0 & 37.9 & 42.5 & 20.5 & 26.0
        & 9.6 & 19.5
        & 28.6 \\

        Cosmos 3 Nano~\cite{cosmos3}\rule[-4pt]{0pt}{13pt}
        & 36.2 & 40.5 & \textbf{46.0} & 40.9
        & 76.8 & 61.3 & 31.8
        & 27.0 & 39.5 & 42.3 & 17.5 & 40.5
        & 20.0 & 16.8
        & 37.3 \\

        \quad + Interleaved Thinking\rule[-4pt]{0pt}{13pt}
        & 30.1 & 36.4 & 41.4 & 36.0
        & 73.7 & 53.1 & 23.9
        & 32.0 & 28.4 & 39.1 & 21.7 & 40.0
        & 4.5 & 19.0
        & 33.5 \\

        % Our configurations: black text and purple highlighting
        \arrayrulecolor{rynn}
        \specialrule{0.9pt}{3pt}{0pt}

        \rowcolor{rynn!8}
        \geucell{\rule[-0.5ex]{0pt}{2.8ex}\textbf{RynnUnify}}\rule[-4pt]{0pt}{13pt}
        & \geucell{35.9}
        & \geucell{44.2}
        & \geucell{43.8}
        & \geucell{41.3}
        & \geucell{78.9}
        & \geucell{\underline{65.7}}
        & \geucell{35.9}
        & \geucell{\textbf{38.5}}
        & \geucell{\underline{\textbf{52.9}}\rlap{$^{\dagger}$}}
        & \geucell{\textbf{57.0}}
        & \geucell{22.0}
        & \geucell{57.5}
        & \geucell{22.6}
        & \geucell{8.5}
        & \geucell{44.0} \\

        \rowcolor{rynn!8}
        \geucell{\rule[-0.5ex]{0pt}{2.8ex}\quad\textbf{+ Agentic Thinking}}\rule[-4pt]{0pt}{13pt}
        & \geucell{\textbf{36.3}}
        & \geucell{\textbf{52.6}}
        & \geucell{44.8}
        & \geucell{\textbf{44.6}}
        & \geucell{\textbf{81.5}\rlap{$^{\dagger}$}}
        & \geucell{\textbf{67.0}\rlap{$^{\dagger}$}}
        & \geucell{\textbf{66.2}\rlap{$^{\dagger}$}}
        & \geucell{36.0}
        & \geucell{51.8}
        & \geucell{56.5}
        & \geucell{\textbf{25.5}}
        & \geucell{\textbf{61.0}}
        & \geucell{\textbf{25.0}\rlap{$^{\dagger}$}}
        & \geucell{\textbf{29.1}\rlap{$^{\dagger}$}}
        & \geucell{\textbf{50.0}\rlap{$^{\dagger}$}} \\

        \specialrule{0.9pt}{0pt}{0pt}
        \arrayrulecolor{black}

    \end{tabularx}
\end{table*}

% Preamble

% Black text with a slightly larger font for our row.
\providecommand{\vpcell}[1]{#1}
\renewcommand{\vpcell}[1]{{\normalsize\textcolor{black}{#1}}}

\subsection{Embodied Understanding Capability}

\subsubsection{Foundational Embodied Understanding}

\paragraph{Evaluation protocol.}
We evaluate embodied cognition on VSI-Bench~\cite{yang2025thinking}, MMSI-Bench~\cite{yang2025mmsi}, MindCube~\cite{wang2025mindcube}, ERQA~\cite{team2025gemini}, RoboSpatial~\cite{song2025robospatial}, and EmbSpatial~\cite{du2024embspatial}, and localization on RefSpatial-Bench~\cite{zhou2025roborefer}, ShareRobot (Affordance and Trajectory)~\cite{robobrain}, and Where2Place~\cite{yuan2024robopoint} in Table~\ref{tab:embodied_understanding}. All methods answer directly from benchmark inputs. RynnUnify uses only its foundational reasoner, without invoking the generator or the additional RL-trained agent.
We report benchmark-specific metrics and their unweighted mean after 0–100 normalization.

\paragraph{Main results.}
RynnUnify achieves the highest average score among the evaluated UMMs (Table~\ref{tab:embodied_understanding}), reaching 57.9 compared with 53.2 for Cosmos 3 Nano and leading on six of ten metrics. It also compares favorably with general-purpose multimodal models, demonstrating strong embodied cognition and localization capabilities across spatial understanding, interaction grounding, and trajectory prediction. On VSI-Bench and RoboSpatial, it outperforms Cosmos 3 Nano by 12.2 and 10.7 points, respectively, showing stronger spatial understanding across complementary settings. For localization, ShareRobot Affordance improves from 28.0 to 41.8, indicating more accurate grounding of task-relevant interaction points. RynnUnify also achieves the best results across all compared models on EmbSpatial (83.0), RefSpatial-Bench (60.3), and ShareRobot Trajectory (0.27, lower is better). These complementary capabilities provide a foundation for interleaved video--language reasoning: understanding scene geometry and interaction targets supports subtask planning and world-model prompt formulation, while spatial and temporal reasoning helps interpret the outcomes of subsequent video predictions.
Additional embodied cognition and localization examples are provided in Figures~\ref{fig:more-cognition-cases} and~\ref{fig:more-localization-cases} in the appendix.

\begin{figure*}[t]
    \centering
    \includegraphics[width=\textwidth]{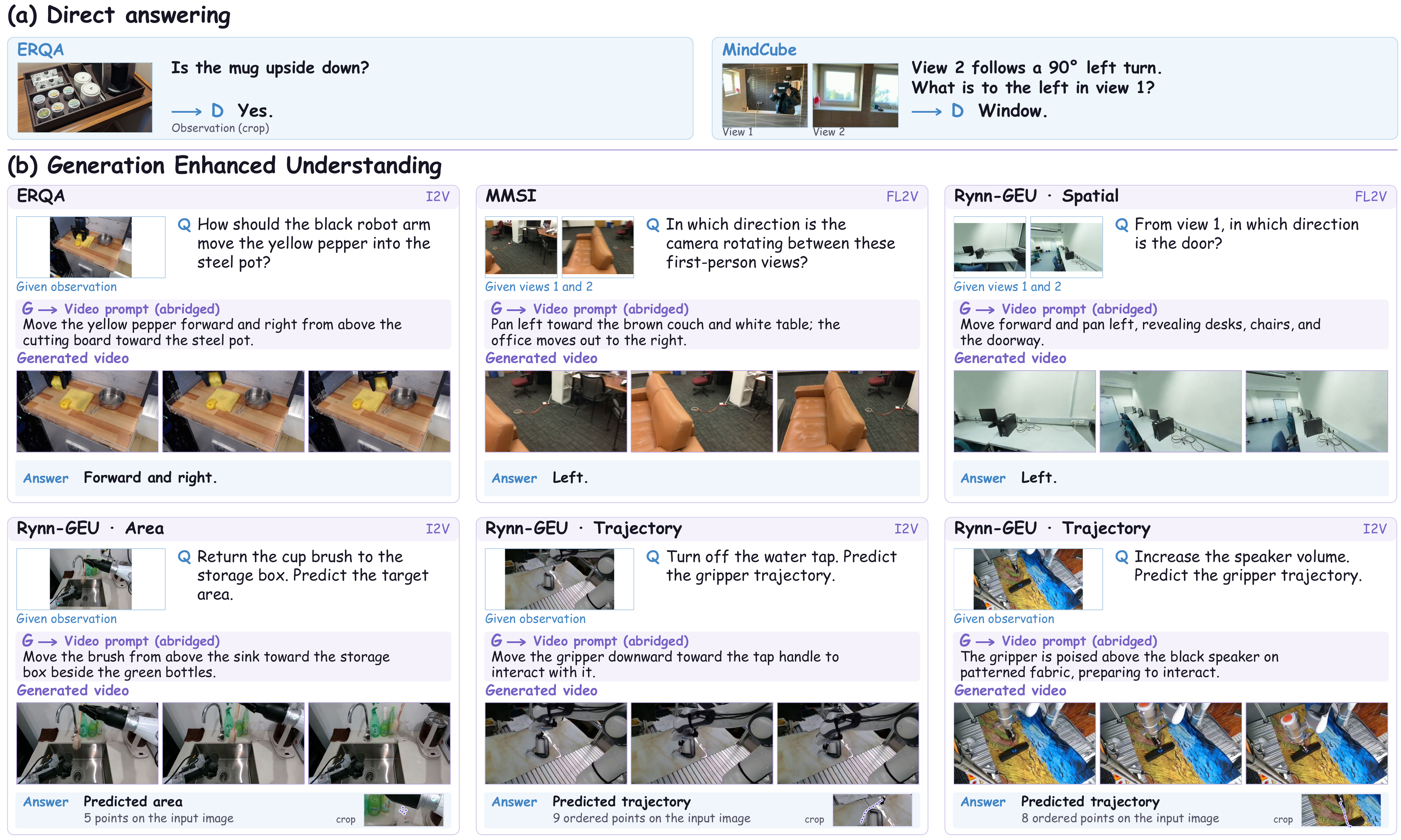}

% [original]
%     \caption{.}

% [revised]
    \caption{\textbf{Qualitative examples of direct and generation-assisted embodied reasoning.} (a) Direct answering on ERQA and MindCube using the input observations. (b) Generation-assisted reasoning on ERQA, MMSI-Bench, and Rynn-GEU. Each generation-assisted example shows the input, question, an abridged video prompt, sampled frames from the generated video, and the final answer. I2V and FL2V denote image-to-video and first-and-last-frame-to-video generation, respectively. Predicted area points and trajectories are overlaid on crops of the input image.}

    \label{fig:geu-cases}
\end{figure*}
\subsubsection{Embodied Reasoning with Agentic Visual Generation}

\paragraph{Main results.}
Table~\ref{tab:geu_main_results} evaluates embodied reasoning both without explicit video generation and with generated visual context. On Rynn-GEU, RynnUnify achieves the strongest direct-answering performance among the compared UMMs, with an average score of 44.0. This highlights its intrinsic capacity for anticipatory embodied reasoning: inferring action consequences, physical changes, and spatial relationships beyond what is directly observable in the input. These results support the motivation for our foundational embodied pretraining, which develops the spatial, physical, and action-related knowledge needed to reason about unobserved states. This predictive understanding also provides a basis for formulating task-relevant video prompts and interpreting generated visual evidence.

% [original]
% The agentic extension further increases the Rynn-GEU average to 50.0, the highest among all compared methods, exceeding GPT-5.6-Terra by 4.9 points. It improves seven of ten tasks over direct answering and outperforms the other UMM rollout variants on all ten. The gains span perception-related grounding, spatial reasoning, and physical prediction, with particularly large improvements on Area (35.9 to 66.2) and Mess (8.5 to 29.1). Improvements also extend to the public reasoning benchmarks, where the average rises from 41.6 to 44.2, including an 8.4-point gain on MindCube. Together, these results demonstrate the value of combining a strong embodied reasoning foundation with agentic use of video prediction.

% [revised]
The agentic extension further increases the Rynn-GEU average to 50.0, the highest among all compared methods, exceeding GPT-5.6-Terra by 4.9 points. It improves seven of ten tasks over direct answering and outperforms the other UMM rollout variants on all ten. The gains span perception-related grounding, spatial reasoning, and physical prediction, with particularly large improvements on Area (35.9 to 66.2) and Mess (8.5 to 29.1). Improvements also extend to all three public reasoning benchmarks, where the average rises from 41.3 to 44.6, including an 8.4-point gain on MindCube. Together, these results demonstrate the value of combining a strong embodied reasoning foundation with agentic use of video prediction.

\paragraph{Effect of generated visual context.}
% [original]
% Explicit video generation does not automatically improve reasoning. On Rynn-GEU, interleaved rollouts reduce the average scores of BAGEL, RxBrain, and Cosmos 3 Nano from 11.9 to 10.8, 31.4 to 28.6, and 37.3 to 33.5, respectively. A plausible explanation is that inaccurate dynamics, inconsistent geometry, or task-irrelevant predictions introduce misleading evidence into subsequent reasoning. We hypothesize that RynnUnify's gains reflect three complementary factors. First, selective generation preserves a direct-answering path instead of requiring a rollout for every question. Second, stronger embodied understanding supports more robust interpretation of imperfect videos, allowing the reasoner to extract useful cues while remaining grounded in the original observations. Third, precise world-model prompt formulation helps focus video prediction on the interactions and state changes relevant to the question, improving the usefulness of generated evidence. Nevertheless, position, distance, and size scores decrease slightly after agentic rollout, suggesting that fine-grained geometric judgments remain sensitive to distortions in predicted scenes.

% [revised]
Generated visual context has different effects across models and benchmarks.
On the public benchmarks, BAGEL's interleaved rollout increases its average from 35.1 to 38.4, with MindCube improving from 34.7 to 39.6 and ERQA from 39.5 to 44.8, while MMSI-Bench decreases slightly from 31.0 to 30.7.
RxBrain instead declines on all three benchmarks, from 35.8 to 31.8 on MMSI-Bench, 47.5 to 45.3 on MindCube, and 42.8 to 37.0 on ERQA, reducing its average from 42.0 to 38.0.
RynnUnify's agentic rollout improves all three scores and reaches the highest public-benchmark average among the evaluated UMMs and their reasoning variants, at 44.6 compared with 38.4 for BAGEL and 38.0 for RxBrain.
On ERQA, RynnUnify and BAGEL both score 44.8 at the reported precision; RynnUnify's stronger aggregate performance reflects higher MMSI-Bench and MindCube scores.

On Rynn-GEU, interleaved rollouts reduce the average scores of BAGEL, RxBrain, and Cosmos 3 Nano from 11.9 to 10.8, 31.4 to 28.6, and 37.3 to 33.5, respectively.
Together with BAGEL's gains on the public benchmarks, these results indicate that the usefulness of generated visual context depends on the model and the task.
A plausible explanation for the declines is that inaccurate dynamics, inconsistent geometry, or task-irrelevant predictions introduce misleading evidence into subsequent reasoning.
We hypothesize that RynnUnify's gains reflect three complementary factors.
First, selective generation preserves a direct-answering path instead of requiring a rollout for every question.
Second, stronger embodied understanding supports more robust interpretation of imperfect videos, allowing the reasoner to extract useful cues while remaining grounded in the original observations.
Third, precise world-model prompt formulation helps focus video prediction on the interactions and state changes relevant to the question, improving the usefulness of generated evidence.
Nevertheless, position, distance, and size scores decrease slightly after agentic rollout, suggesting that fine-grained geometric judgments remain sensitive to distortions in predicted scenes.

\subsubsection{Case Study}

Figure~\ref{fig:geu-cases} illustrates direct and generation-assisted reasoning across public benchmarks and Rynn-GEU. In panel (a), the model answers from the input observations, covering object-state recognition in ERQA and viewpoint reasoning in MindCube. In panel (b), the agent formulates a task-specific video prompt, inspects the generated sequence, and returns a textual or spatial answer. The ERQA example relates the motion of a yellow pepper to the location of the target pot. The FL2V examples use transitions between given views to reason about camera rotation in MMSI-Bench and the direction of a doorway in Rynn-GEU.

The Rynn-GEU Area and Trajectory examples connect generated visual context with spatial predictions in the original observation. For brush storage, the agent returns points identifying the target placement region. For tap control and speaker-volume adjustment, it returns ordered gripper trajectories. The predicted points and paths are overlaid on crops of the input image, linking the anticipated interaction to locations in the observed scene. Together, these examples illustrate how task-focused video generation supplies intermediate context for action reasoning, viewpoint reasoning, and spatial grounding, while retaining a direct-answering path.
Additional generation-assisted reasoning examples are provided in Figure~\ref{fig:more-agentic-cases} in the appendix.

\begin{table}[t]
    \centering
    \caption{
        Long-horizon visual planning and generation quality on Rynn-VP.
        We report Planning Intelligence (PI), Visual Realization (VR),
        and Visual Quality (VQ), all on a 0--100 scale.
        Overall is computed as
        $0.50 \times \mathrm{PI} + 0.40 \times \mathrm{VR}
        + 0.10 \times \mathrm{VQ}$.
        \textbf{Bold} indicates the best result among UMMs.
        Higher is better for all metrics.
    }
    \label{tab:rynn_vp_results}

    %\footnotesize
    \color{black}
    \setlength{\tabcolsep}{4pt}
    \renewcommand{\arraystretch}{1.2}

    \begin{tabularx}{\columnwidth}{
        >{\hsize=2.5\hsize\linewidth=\hsize\raggedright\arraybackslash}X
        >{\hsize=1.1\hsize\linewidth=\hsize\centering\arraybackslash}X
        *{4}{>{\hsize=0.6\hsize\linewidth=\hsize\centering\arraybackslash}X}
    }
        \toprule
        \textbf{Method}
        & \textbf{Output}
        & \textbf{PI}
        & \textbf{VR}
        & \textbf{VQ}
        & \textbf{Overall} \\
        \midrule

        \rowcolor{black!5}
        \multicolumn{6}{l}{
            \textbf{Composed Video Generation System}
        } \\

        GPT-5.6 Terra + MiniMax-H3
        & Video
        & 59.82
        & 69.56
        & 79.73
        & 65.71 \\

        \midrule

        \rowcolor{black!5}
        \multicolumn{6}{l}{
            \textbf{Unified Multimodal Models}
        } \\

        BAGEL
        & Step images
        & 9.46
        & 24.36
        & 58.50
        & 20.32 \\

        RxBrain
        & Step images
        & 1.28
        & 11.35
        & 28.15
        & 8.00 \\

        Cosmos 3 Nano
        & Video
        & 30.50
        & 54.58
        & \textbf{79.69}
        & 45.05 \\

        \arrayrulecolor{rynn}
        \specialrule{0.7pt}{2pt}{0pt}

        \rowcolor{rynn!8}
        \textbf{RynnUnify}
        & Video
        & \textbf{45.66}
        & \textbf{60.54}
        & 77.53
        & \textbf{54.80} \\

        \specialrule{0.7pt}{0pt}{0pt}
        \arrayrulecolor{black}
    \end{tabularx}
\end{table}

\subsection{Interleaved Visual Planning}
\begin{figure*}[t]
    \centering
    \includegraphics[width=\textwidth]{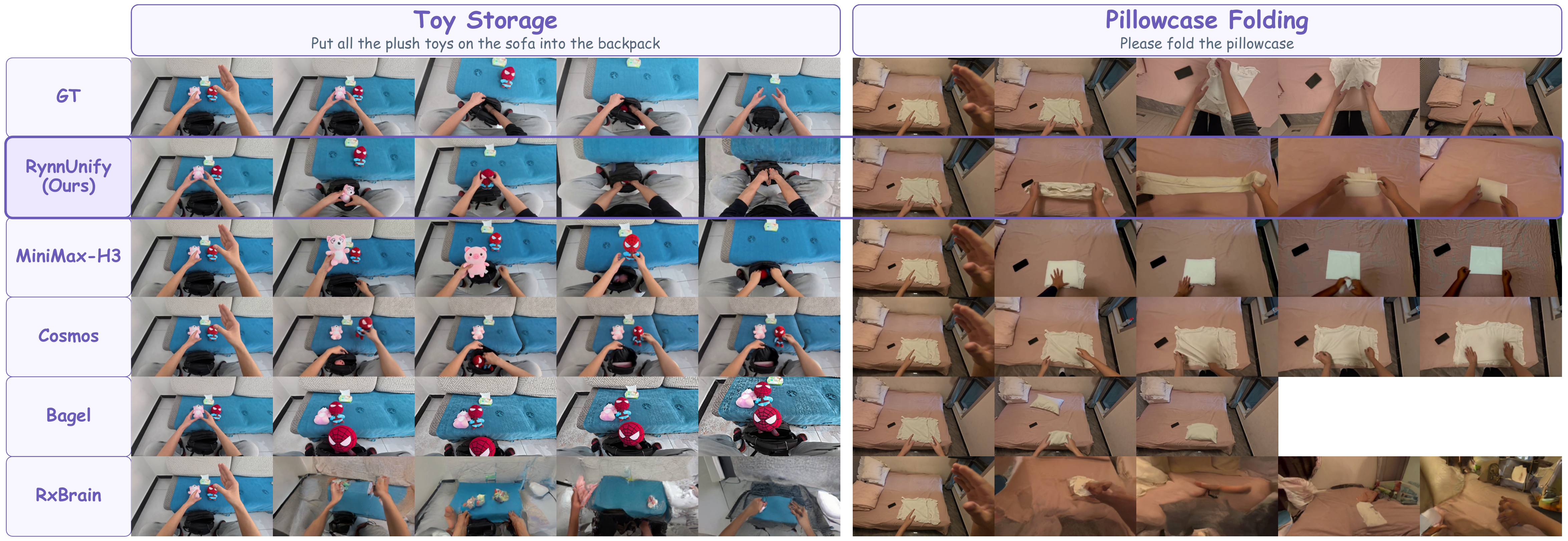}
    \caption{Visual planning comparisons on toy storage and pillowcase folding.
    RynnUnify (Ours) is highlighted in purple.}
    \label{fig:visual-planning-cases}
\end{figure*}

\subsubsection{Evaluation Settings}

% [original]
% \paragraph{Evaluation protocol.}
% Evaluation is conducted on Rynn-VP, which contains 200 egocentric task scenarios, each initialized with a scene image and a high-level task instruction. The interleaved planning setting alternates between subtask reasoning and visual generation, using predicted scene states to inform subsequent planning. Rollouts terminate when the model indicates task completion or reaches a maximum of 15 steps.
% We follow the scoring protocol in Section~\ref{sec:rynn_vp_benchmark} and report PI, VR, VQ, and the overall score. Reference videos and subtask annotations are used only for evaluation. Videos are generated at 480p and 24 fps, and results are averaged over three evaluation runs. For video generation models, we use a default duration of 4 seconds per step. We enable the dynamic duration prediction mechanism of RynnUnify to adapt each generated segment to the planned subtask.

% [revised]
\paragraph{Evaluation protocol.}
Evaluation is conducted on Rynn-VP, which contains 200 egocentric task scenarios, each initialized with a scene image and a high-level task instruction. The interleaved planning setting alternates between subtask reasoning and visual generation, using predicted scene states to inform subsequent planning. Rollouts terminate when the model indicates task completion or reaches a maximum of 15 steps.
We follow the scoring protocol in Section~\ref{sec:rynn_vp_benchmark} and report PI, VR, VQ, and the overall score. Reference videos and subtask annotations are used only for evaluation. Videos are generated at 480p and 24 fps. For video generation models, we use a default duration of 4 seconds per step. We enable the dynamic duration prediction mechanism of RynnUnify to adapt each generated segment to the planned subtask.

\paragraph{Compared methods.}
We compare RynnUnify with BAGEL~\cite{deng2025bagel}, RxBrain~\cite{liang2026rxbrain}, Cosmos 3 Nano~\cite{cosmos3}, and a composed system combining GPT-5.6 Terra~\cite{openai2026gpt56terra} with MiniMax-H3~\cite{minimax2026h3}. In our evaluation, BAGEL and RxBrain produce ordered step images, whereas the other methods generate videos.

\subsubsection{Main Results}

% [original]
% RynnUnify achieves the highest PI, VR, and overall scores among the evaluated UMMs (Table~\ref{tab:rynn_vp_results}), improving the overall score from 45.05 to 54.80 over Cosmos 3 Nano. The PI improvement (+15.16) reflects more effective progress toward task goals, consistent with stronger reasoner support for coordinating interdependent operations. The VR improvement (+5.96) suggests more coherent and physically plausible visual realization of these operations. Scene memory may help preserve task-relevant entities and states across segments, while dynamic duration adapts each prediction to the temporal requirements of the sub-plan. These improvements accompany a slightly lower VQ score (77.53 vs. 79.69), suggesting that the gains extend beyond perceptual quality to goal-directed planning and task-consistent visual prediction.

% [revised]
RynnUnify achieves the highest PI, VR, and overall scores among the evaluated UMMs (Table~\ref{tab:rynn_vp_results}), improving the overall score from 45.05 to 54.80 over Cosmos 3 Nano.
PI increases from 30.50 to 45.66 (+15.16), indicating stronger progress toward task goals and coordination of interdependent operations.
VR increases from 54.58 to 60.54 (+5.96), indicating more consistent and physically plausible visual realization.
These gains accompany a slightly lower VQ score (77.53 vs. 79.69), showing that the overall improvement is driven by planning and visual realization.

Relative to the Cosmos3 backbone, RynnUnify combines embodied fine-tuning with cross-chunk scene memory, self-forcing training, and subtask-conditioned duration prediction.
Scene memory retains initial and historical scene references together with recent motion context, providing persistent cues for object identity, scene layout, and interaction state across successive clips.
Self-forcing exposes the generator to perturbed observations and cached generated endpoints and histories during training, preparing it to continue from the imperfect predictions encountered during inference.
Duration prediction adapts each generated segment to the temporal requirements of the planned operation.
These extensions target complementary challenges in long-horizon rollout: retaining task-relevant scene information and continuing coherently as prediction errors accumulate.
The stronger PI and VR results are consistent with this design goal, demonstrating improved task progression and visual consistency at the system level.

\subsubsection{Visualization}
% 正文：双栏论文跨两栏
Figure~\ref{fig:visual-planning-cases} presents qualitative comparisons on Toy Storage and Pillowcase Folding, illustrating task progression and visual consistency across successive predictions. RynnUnify interleaves subtask reasoning with video generation, using the predicted scene to inform subsequent planning. In Toy Storage, it progresses from manipulating one toy to collecting subsequent toys, while maintaining the backpack and sofa as consistent spatial references. In Pillowcase Folding, it generates successive fabric manipulations that progressively produce a compact folded configuration. These examples cover complementary challenges: tracking multiple objects across repeated transfers and maintaining coherent object transformations during deformable-object manipulation.

The baseline sequences exhibit several characteristic difficulties. Some show repeated interactions with limited visible task progression, whereas others introduce abrupt changes in object appearance, scale, or background layout. These differences highlight that plausible individual frames alone are insufficient for long-horizon visual planning: successive predictions must preserve relevant scene state while advancing the task. RynnUnify exhibits more coherent task progression in these examples, illustrating how interleaved reasoning and video prediction can connect individual operations into a sustained plan. The reference demonstrations provide examples of successful execution; effective predictions need not reproduce their exact motions or intermediate configurations.
Additional long-horizon visual planning cases are provided in Figure~\ref{fig:more-vp-cases} in the appendix.

\subsection{Evaluation of the Post-trained WAM}

\subsubsection{Evaluation Settings on Simulation Benchmarks}
\label{sec:simulation_settings}

\paragraph{Benchmark and evaluation protocol.}
We evaluate RynnUnify on the Composite-Seen and Composite-Unseen splits of RoboCasa365~\cite{robocasa365}. We also report results on 20 composite mobile-manipulation tasks, referred to as Selected-20, including \texttt{PackIdenticalLunches}, \texttt{SeparateFreezerRack}, and \texttt{DeliverStraw}. These tasks involve base movement and multiple manipulation steps under changing camera views. We run 10 episodes per task on Selected-20, for a total of 200 episodes, and report task success rates. Results for the full composite splits and Selected-20 are given in Table~\ref{tab:robocasa_results}.

\paragraph{Training and policy configuration.}
We initialize our WAM from pretrained RynnUnify and use approximately 1,600 hours of atomic-task trajectories and 177 hours of composite-task trajectories from RoboCasa365 for robot adaptation. As described in Section~\ref{sec:wam_training}, we first train the reasoner with subtask supervision while keeping the generator frozen. We then freeze the reasoner and train the generator with video and native-action supervision. Finally, we jointly train the generator and a separate action head, with auxiliary camera-pose supervision.

At each decision step, the reasoner receives the task instruction and at most 12 RGB images in total. These include the current left, wrist, and right camera views, one left-view image from the beginning of the episode, up to five earlier left-view images, and up to three earlier wrist images. Historical images are selected at fixed time offsets. The generator uses the current observations and the subtask latent from the reasoner. The action head predicts a $48\times12$ action chunk, of which the first 16 actions are used for execution. The 12-dimensional action space includes planar base translation, base yaw, torso height, end-effector translation and rotation, gripper control, and a control-mode indicator.

% \paragraph{Compared methods.}
% We compare RynnUnify with $\pi_{0}$~\cite{pi0}, $\pi_{0.5}$~\cite{pi_0.5}, GR00T N1.5~\cite{nvidia2025grootn15}, GR00T N1.6~\cite{nvidia2025grootn16}, Cosmos3~\cite{cosmos3}, Xiaomi-Robotics-1~\cite{xiaomi2026robotics1}, and CMA~\cite{wang2026crackingmobilemanipulationfailure}, with results summarized in Table~\ref{tab:robocasa_results}.

\paragraph{Results on RoboCasa365.}
% As shown in Table~\ref{tab:robocasa_results}, RynnUnify achieves success rates of 52.0\% on Composite-Seen and 50.0\% on Composite-Unseen, ranking second on the former and first among the compared methods on the latter. On Composite-Unseen, it exceeds CMA and Cosmos3 by 6.0 and 11.0 percentage points, respectively. On Selected-20, RynnUnify achieves the highest success rate of 51.5\%, outperforming CMA, Xiaomi-Robotics-1, and Cosmos3 by 4.5, 5.5, and 6.5 percentage points, respectively. These selected mobile-manipulation tasks require coordinating movement with successive manipulation operations while maintaining task-relevant scene understanding under changing viewpoints. We hypothesize that the generator's pretraining on egocentric videos with viewpoint changes contributes to this capability by encouraging it to model scene evolution under camera motion. Joint video-action training connects these predictive representations with action generation, supporting coordinated robot repositioning and object manipulation.

As shown in Table~\ref{tab:robocasa_results}, RynnUnify achieves success rates of 52.0\% on Composite-Seen and 50.0\% on Composite-Unseen, ranking second and first, respectively, among the compared methods. On Composite-Unseen, it exceeds CMA and Cosmos3 by 6.0 and 11.0 percentage points. On Selected-20, our WAM achieves the highest success rate of 51.5\%, exceeding CMA, Xiaomi-Robotics-1, and Cosmos3 by 4.5, 5.5, and 6.5 percentage points, respectively. These results are relevant to mobile manipulation, where the policy must coordinate base movement and object manipulation as its viewpoint changes.

% Preamble

\definecolor{rynn}{RGB}{108,92,186}
\begin{table}[t]
\centering
\footnotesize
\setlength{\tabcolsep}{3pt}
\renewcommand{\arraystretch}{1.12}
\caption{\textbf{Comparison on RoboCasa365.}
Success rates (\%) on Composite-Seen, Composite-Unseen, and Selected-20. Selected-20 consists of 20 composite mobile-manipulation tasks, with 10 trials per task. Baseline results on the full composite splits are taken from published reports, while Selected-20 results are obtained from our evaluation. \textbf{Bold} indicates the best result in each column.}
\label{tab:robocasa_results}
\begin{tabularx}{\columnwidth}{@{}l*{3}{>{\centering\arraybackslash}X}@{}}
\toprule
\textbf{Method}
& \shortstack{\textbf{Composite Seen}}
& \shortstack{\textbf{Composite Unseen}}
& \shortstack{\textbf{Selected 20 tasks}} \\
\midrule
% \rowcolor{rynn!8}

$\pi_{0}$~\cite{pi0}            & 6.1  & 1.1  & 3.0  \\
$\pi_{0.5}$~\cite{pi_0.5}          & 7.1  & 1.2  & 2.0  \\
GR00T N1.5~\cite{nvidia2025grootn15}           & 14.8 & 2.7  & 7.5  \\
GR00T N1.6~\cite{nvidia2025grootn16}            & 9.4  & 1.7  & 6.5  \\

Cosmos 3~\cite{cosmos3} & 49.6 & 39.0 & 45.0 \\

Xiaomi-Robotics-1~\cite{xiaomi2026robotics1} & \textbf{57.1} & 32.1 & 46.0 \\

CMA \cite{wang2026crackingmobilemanipulationfailure} & 50.6 & 44.0 & 47.0 \\

\arrayrulecolor{rynn}
\specialrule{0.7pt}{2pt}{0pt}
\rowcolor{rynn!8}
\textbf{RynnUnify}
& 52.0 & \textbf{50.0} & \textbf{51.5} \\
\specialrule{0.7pt}{0pt}{0pt}
\arrayrulecolor{black}
\end{tabularx}
\end{table}

\begin{table*}[t]
    \centering
    \caption{
        Stage-wise and task success rates (\%) on Astribot S1 over 20 trials per task.
        Stage success is scored separately for each operation, regardless of success in preceding stages.
        Task SR measures complete-task success, and Avg.\ SR averages the three task success rates.
        \textbf{Bold} denotes column-wise best results, including ties.
    }
    \label{tab:real_robot_results}

    \begingroup
    \footnotesize
    \color{black}
    \arrayrulecolor{black}
    \setlength{\tabcolsep}{2.0pt}
    \renewcommand{\arraystretch}{1.18}

    \begin{tabularx}{\textwidth}{@{}l
        *{4}{>{\centering\arraybackslash}X}
        @{\hspace{6pt}}
        *{4}{>{\centering\arraybackslash}X}
        @{\hspace{6pt}}
        *{4}{>{\centering\arraybackslash}X}
        @{\hspace{6pt}}
        >{\centering\arraybackslash}X@{}}
        \toprule
        \multirow{2}{*}[-2.4ex]{\textbf{Model}}
        & \multicolumn{4}{c}{\textbf{Turn-and-Grasp}}
        & \multicolumn{4}{c}{\textbf{Block Swapping}}
        & \multicolumn{4}{c}{\textbf{Bowl $\times 3$ then Spoon}}
        & \multirow{2}{*}[-2.4ex]{\makecell{\textbf{Avg.}\\\textbf{SR}}} \\
        \cmidrule(lr){2-5}
        \cmidrule(lr){6-9}
        \cmidrule(lr){10-13}
        \addlinespace[1pt]
        & \makebox[0pt][c]{\makecell{Grasp\\Block 1}}
        & \makebox[0pt][c]{\makecell{Place\\Block 1}}
        & \makebox[0pt][c]{\makecell{Grasp\\Block 2}}
        & \makebox[0pt][c]{\makecell{\textbf{Task}\\\textbf{SR}}}
        & \makebox[0pt][c]{\makecell{Grasp\\Block 1}}
        & \makebox[0pt][c]{\makecell{Place\\Block 1}}
        & \makebox[0pt][c]{\makecell{Grasp\\Block 2}}
        & \makebox[0pt][c]{\makecell{\textbf{Task}\\\textbf{SR}}}
        & \makebox[0pt][c]{\makecell{Grasp\\Bowl}}
        & \makebox[0pt][c]{\makecell{Three\\Cycles}}
        & \makebox[0pt][c]{\makecell{Grasp\\Spoon}}
        & \makebox[0pt][c]{\makecell{\textbf{Task}\\\textbf{SR}}}
        & \\[1.5pt]
        \midrule
        $\pi_{0.5}$~\cite{pi_0.5}
        & 50.0 & 30.0 & 30.0 & 20.0 & 70.0 & 45.0 & 35.0 & 15.0 & 80.0 & \textbf{60.0} & 50.0 & 40.0 & 25.0 \\

        GR00T N1.7 ~\cite{nvidia2026gr00tn17}
        & 55.0 & 35.0 & 35.0 & 25.0 & 65.0 & 55.0 & 30.0 & 20.0 & \textbf{90.0} & 55.0 & \textbf{60.0} & 35.0 & 26.7 \\

        \arrayrulecolor{rynn}
        \specialrule{0.7pt}{2pt}{0pt}
        \rowcolor{rynn!8}
        \textbf{RynnUnify}
        & \textbf{90.0} & \textbf{75.0} & \textbf{70.0} & \textbf{55.0} & \textbf{85.0} & \textbf{85.0} & \textbf{60.0} & \textbf{35.0} & 80.0 & \textbf{60.0} & \textbf{60.0} & \textbf{50.0} & \textbf{46.7} \\

        \specialrule{0.7pt}{0pt}{0pt}
        \arrayrulecolor{black}
    \end{tabularx}
    \endgroup
\end{table*}
\FloatBarrier
\subsubsection{Evaluation Settings on Real Robot}
\label{sec:real_robot_settings}

\paragraph{Experimental setup.}

% [original]
% We conduct real-world experiments on an Astribot S1 dual-arm mobile robot in indoor tabletop environments. We select three representative tasks to assess mobile manipulation, precise object rearrangement, and execution that depends on interaction history.

\begin{figure*}[t]
    \centering
    \includegraphics[width=\textwidth]{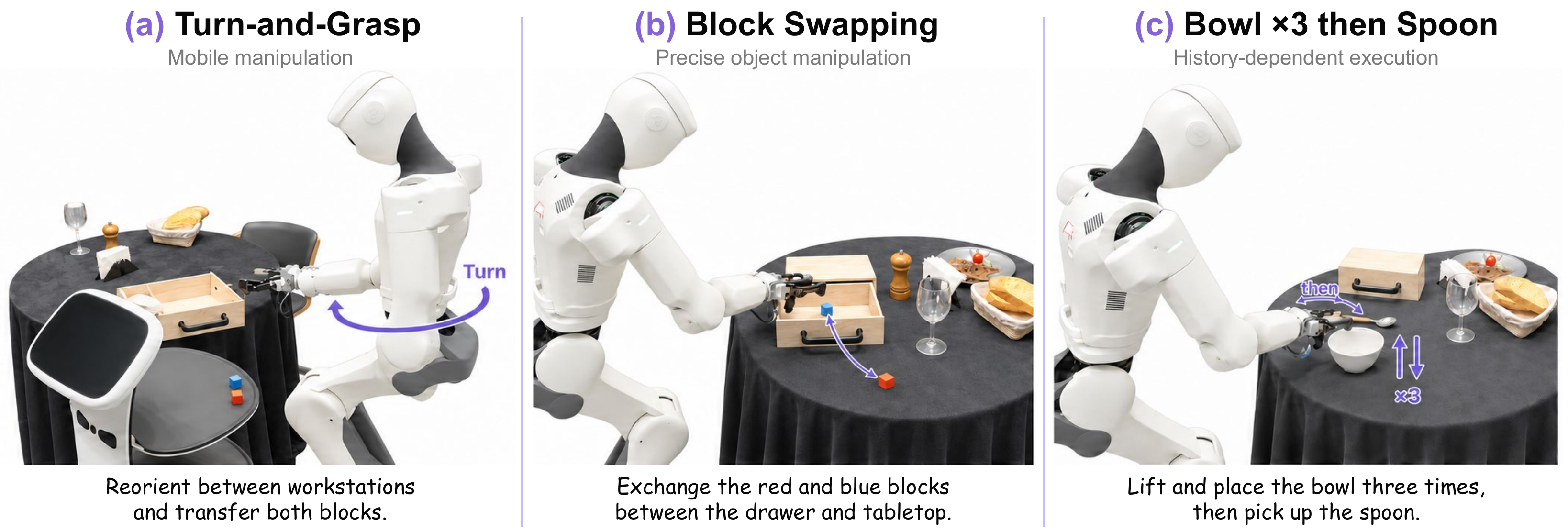}

% [original]
% \caption{Real-robot execution setups of RynnUnify-VLA on Astra S1. }

% [revised]
\caption{\textbf{Real-robot evaluation setups on Astribot S1.} (a) Turn-and-Grasp requires reorientation between workstations to transfer two blocks. (b) Block Swapping exchanges the locations of two blocks between a drawer and the tabletop. (c) Bowl $\times 3$ then Spoon requires three bowl pick-up-and-put-down cycles followed by spoon grasping.}

\label{fig:real_robot_setups}
\end{figure*}

% [revised]
We conduct real-world experiments on an Astribot S1 dual-arm mobile robot in indoor tabletop environments, with the experimental setups shown in Figure~\ref{fig:real_robot_setups}. We select three representative tasks to assess mobile manipulation, precise object rearrangement, and execution that depends on interaction history.

\paragraph{Tasks.}
\textit{(1) Turn-and-Grasp.}
The robot turns toward the target objects and grasps them. This task evaluates coordination between body reorientation and arm manipulation, requiring the model to localize and interact with objects as its viewpoint changes.
\textit{(2) Block Swapping.}
The robot exchanges the positions of a red cube and a blue cube. This task requires identifying the two objects, tracking their locations during manipulation, and placing each cube at the other's initial position. The small size of the cubes places additional demands on visual localization and grasping and placement precision.
\textit{(3) Bowl $\times 3$ then Spoon.}
The robot picks up and puts down a bowl three times before picking up a spoon. This task evaluates whether the model can track completed action cycles and transition to the next subtask at the correct stage. Since successive cycles can yield similar visual observations, correct execution requires tracking task progress through interaction history.

\paragraph{Evaluation protocol.}
We conduct 20 trials per task and report the success rate for each task. A trial is considered successful when the robot completes the full instructed task. For the memory task, success requires exactly three bowl pick-up-and-put-down cycles followed by grasping the spoon in the specified order.

\begin{figure*}[t]
    \centering
    \includegraphics[width=\textwidth]{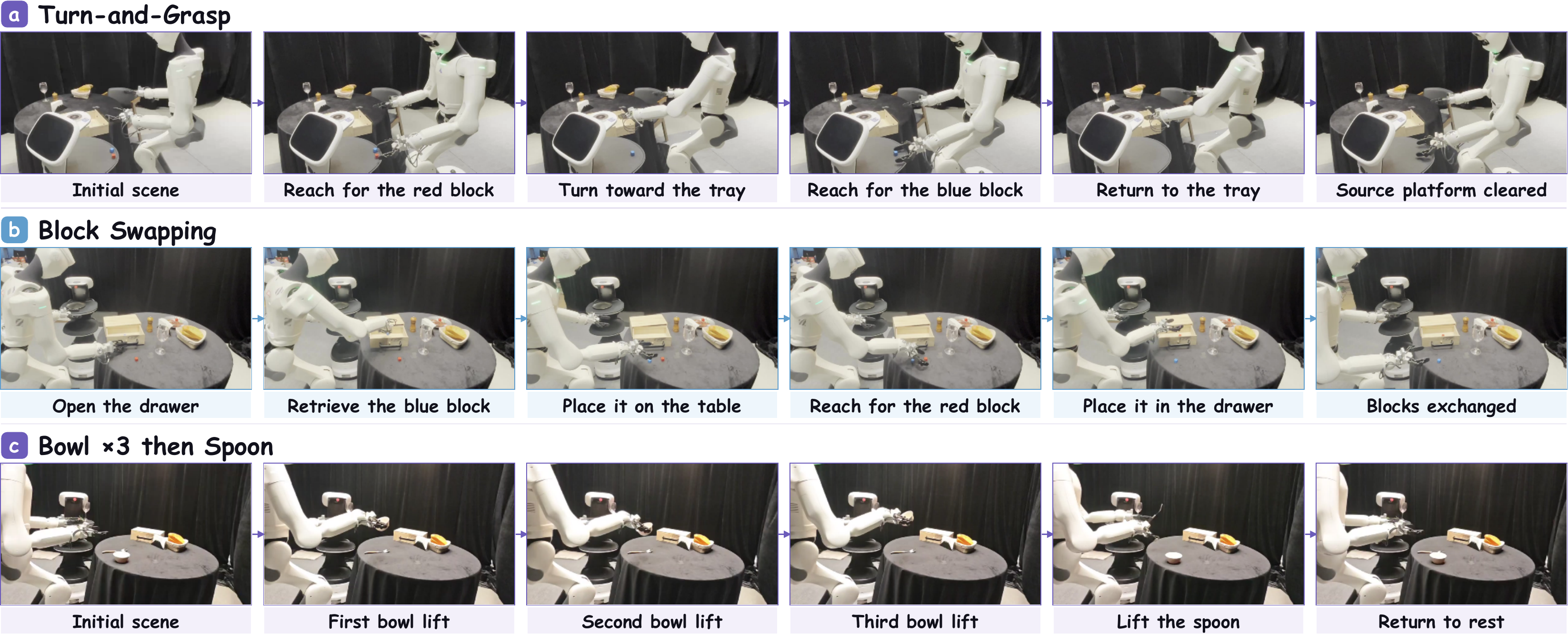}

% [original]
% \caption{Qualitative real-robot execution of RynnUnify-VLA on Astra S1. }

% [revised]
\caption{\textbf{Qualitative real-robot execution of RynnUnify-WAM on Astribot S1.} Rows show successive stages of (a) Turn-and-Grasp, (b) Block Swapping, and (c) Bowl $\times 3$ then Spoon. The sequences illustrate reorientation and block transfer, rearrangement between a drawer and the tabletop, and the transition from repeated bowl manipulation to spoon grasping.}

\label{fig:real_robot}
\end{figure*}

\paragraph{Results on the real robot.}
Figure~\ref{fig:real_robot} shows representative executions of the three tasks, complementing the task-level and stage-wise success rates in Table~\ref{tab:real_robot_results}.
As shown in Table~\ref{tab:real_robot_results}, RynnUnify achieves the highest complete-task success rate on all three tasks, averaging 46.7\%, compared with 26.7\% for GR00T N1.7 and 25.0\% for $\pi_{0.5}$. On Turn-and-Grasp, it achieves 55.0\% success, demonstrating effective coordination between body reorientation and manipulation under changing viewpoints. On Block Swapping, its 35.0\% success rate highlights improved precision in grasping and placing small objects across successive operations. On Bowl $\times 3$ then Spoon, RynnUnify reaches 50.0\% success, exceeding the strongest baseline by 10.0 percentage points. Completing three bowl manipulation cycles before grasping the spoon requires tracking progress across visually similar states, providing behavioral evidence consistent with the utility of visual history for repeated, ordered manipulation.

%% file: sec/7_Conclusion.tex
\section{Conclusion and Future Work}

% [original]
% \textit{[Write this section here.]}

% [revised]
We introduced RynnUnify, a unified embodied model that connects scene understanding, video prediction, and robot action through interleaved video--language reasoning.
Building on Cosmos3, RynnUnify combines embodied supervision with cross-chunk scene memory and self-forcing training to support visual planning across successive operations.
Its agentic extension learns when to invoke video prediction for reasoning about unobserved states, while its WAM extension transfers the model's understanding and predictive representations to action generation through latent planning and joint video--action training.
We also introduced Rynn-VP and Rynn-GEU to evaluate sustained visual planning and anticipatory embodied reasoning.
RynnUnify improves the Rynn-VP overall score from 45.05 to 54.80 over Cosmos 3 Nano, and its agentic extension increases the Rynn-GEU average from 44.0 to 50.0.
Together with results on public embodied understanding benchmarks and simulated and real-robot manipulation, these findings support the value of coordinating reasoning with task-relevant visual predictions.

Several challenges remain.
Generated videos can distort fine-grained geometry, as reflected by the decreases on some spatial tasks after agentic rollout.
Moreover, stronger performance in simulator-free visual planning needs to be complemented by broader validation of physical execution across tasks and environments.
Future work will improve geometric fidelity and the selection of reliable generated evidence, study how visual planning scores relate to robot execution outcomes, and reduce the computational cost of interleaved generation for closed-loop control.

%% file: sec/appendix.tex
\section{Detailed Contributions}

\label{sec:contrib} 
\begin{itemize}
    \item \textbf{Data}: Minghao Zhu, Bohan Hou, Yifei Yang, Kun Xiang,  Yinzhou Tang, Zhikai Wang, Zixuan Wang, Tianyi Zhang, Kehan Li, Xin Li
    \item \textbf{Model Training}:
    \begin{itemize}[label=\raisebox{0.15ex}{\scalebox{0.75}{$\blacksquare$}}]
        \item \textbf{Reasoner Training}: Kehan Li, Xin Li, Bohan Hou, Minghao Zhu, Yifei Yang
        \item \textbf{Generator Training}: Yinzhou Tang, Minghao Zhu, Bohan Hou, Zixuan Wang, Kun Xiang
        \item \textbf{Agentic RL}: Kun Xiang, Minghao Zhu, Yinzhou Tang, Zhikai Wang, Bohan Hou
        \item \textbf{WAM Post-Training}: Zixuan Wang, Bohan Hou, Tianyi Zhang, Minghao Zhu, Yinzhou Tang
    \end{itemize}
    \item \textbf{Infrastructure}: Kun Xiang, Kehan Li, Yinzhou Tang, Minghao Zhu, Zhikai Wang, Tianyi Zhang, Xin Li
    \item \textbf{Evaluation}: Bohan Hou, Yifei Yang, Minghao Zhu, Kun Xiang, Zixuan Wang, Tianyi Zhang, Zhikai Wang
    %\item \textbf{Hardware and Robot System}: Yaxi Zhao, Minghua Zeng, Junlong Gao
    %\item \textbf{Senior Advisory}: Jun Cen, Siteng Huang, Wenqiao Zhang (Zhejiang University), Chengju Liu (Tongji University), Jianfei Yang (Nanyang Technological University), Shijian Lu (Nanyang Technological University), Deli Zhao
\end{itemize}

% [original]
% \subsection{Qualitative Examples}
%
% \textit{[Write this section here.]}
%
% \subsubsection{Examples for Embodied Cognition}
%
% \textit{[Write this section here.]}
%
% \subsubsection{Examples for Embodied Location}
%
% \textit{[Write this section here.]}
%
% \subsubsection{Examples for General Visual Understanding}
%
% \textit{[Write this section here.]}
%
% \subsubsection{Examples for physically Grounded Reasoning}
%
% \textit{[Write this section here.]}
%
% \subsubsection{Examples for Navigation}
%
% \textit{[Write this section here.]}
%
% \subsubsection{Examples for Manipulation Planning}
%
% \textit{[Write this section here.]}

% [revised]
\section{Qualitative Examples}

We provide additional examples of embodied cognition, localization, agentic rollout, and long-horizon visual planning to complement the quantitative evaluations in the main paper.

\subsection{Embodied Cognition}

\begin{figure*}[p]
    \centering
    \includegraphics[width=\textwidth,height=0.88\textheight,keepaspectratio]{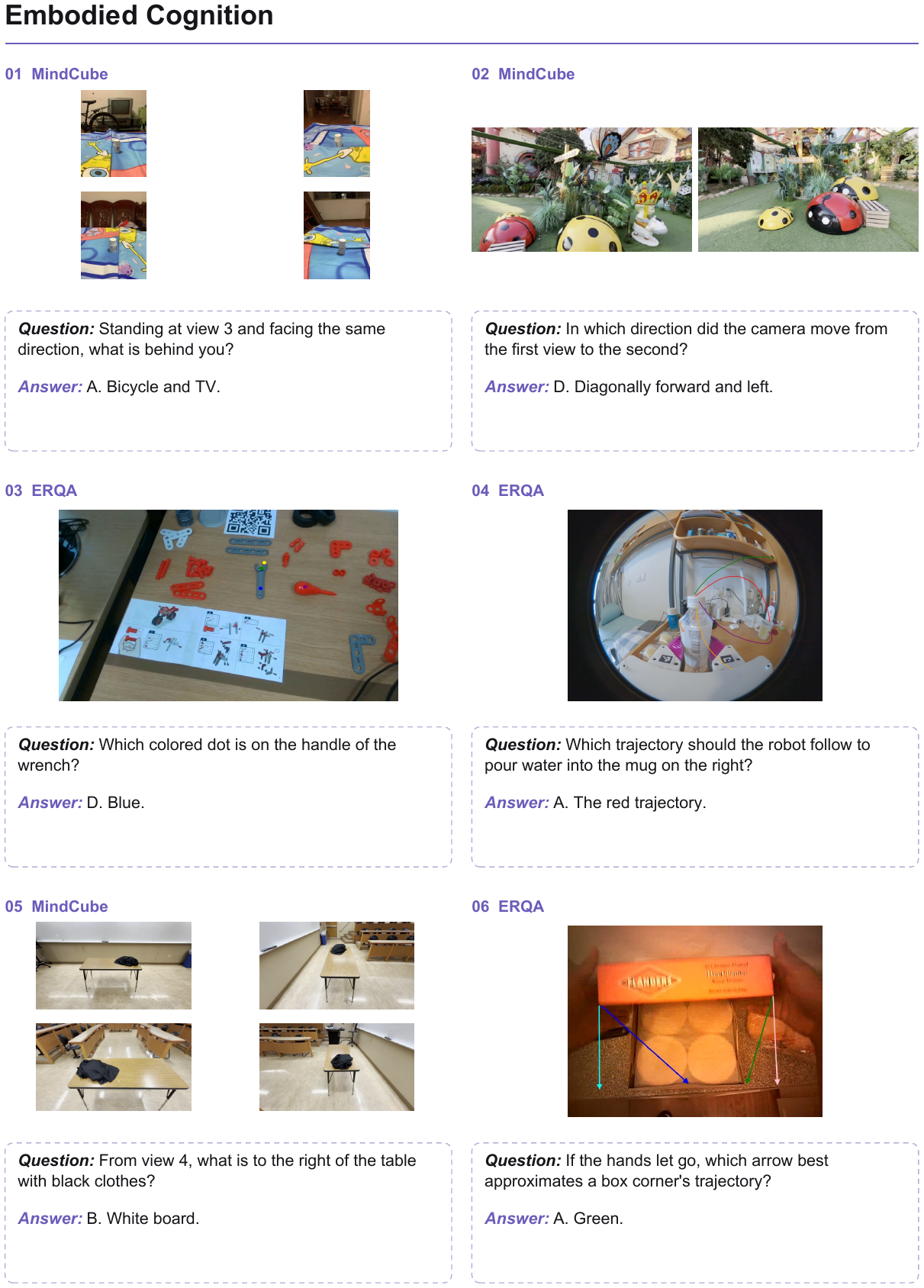}
    \caption{\textbf{Additional embodied cognition examples.} Six MindCube and ERQA cases cover viewpoint reasoning, camera motion, interaction grounding, and anticipated object motion. Each case presents the visual observations, question, and answer.}
    \label{fig:more-cognition-cases}
\end{figure*}

Figure~\ref{fig:more-cognition-cases} presents six examples from MindCube and ERQA. The cases cover reasoning across viewpoints, camera motion, interaction grounding, and anticipated object motion, illustrating how visual observations are connected to task-relevant textual answers.

\subsection{Embodied Localization}

\begin{figure*}[p]
    \centering
    \includegraphics[width=\textwidth,height=0.88\textheight,keepaspectratio]{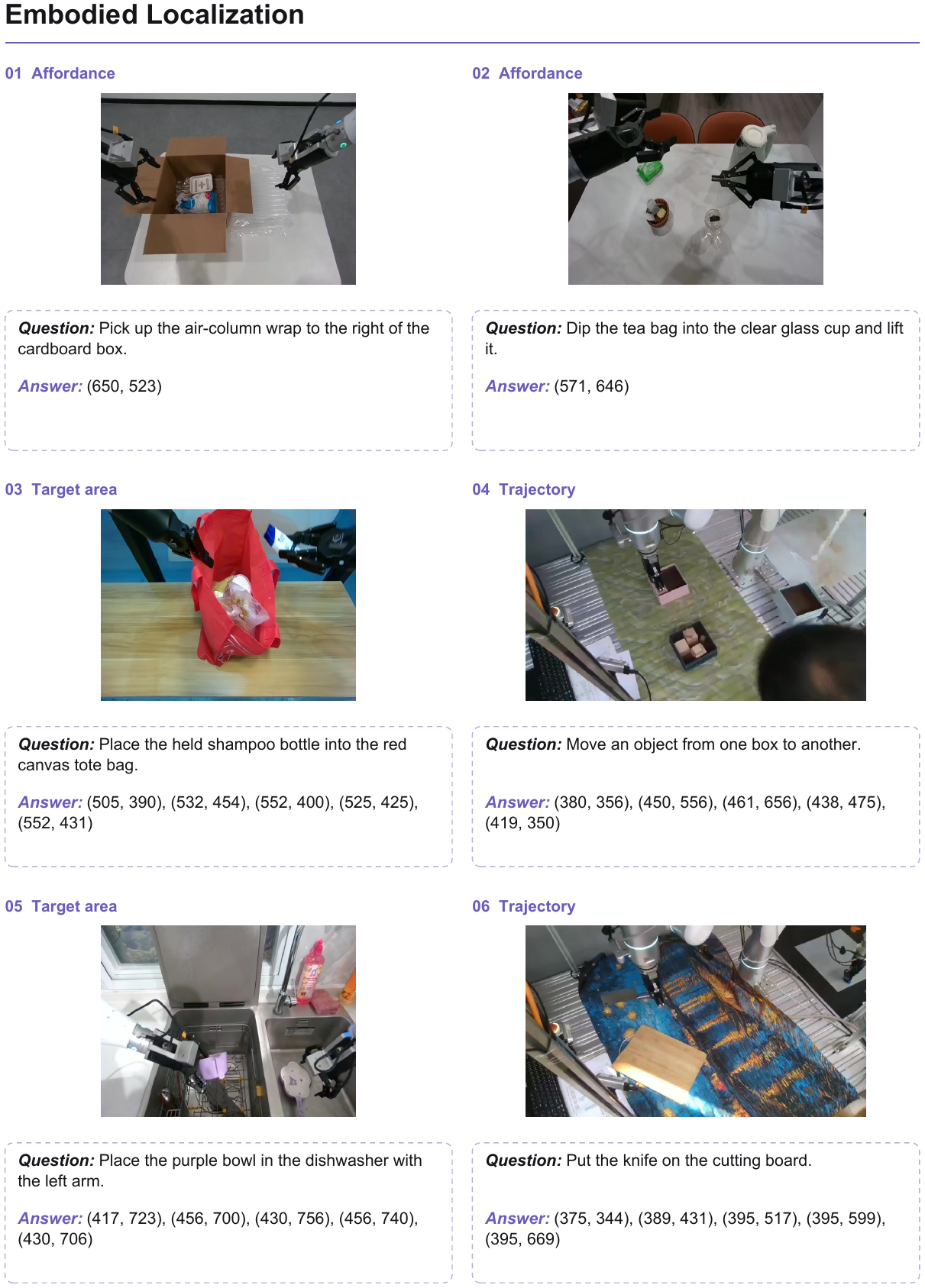}
    \caption{\textbf{Additional embodied localization examples.} The six cases comprise two affordance, two target-area, and two trajectory questions. Answers specify an interaction point, a set of target-region points, or an ordered trajectory, respectively, grounding the task instruction in the input observation.}
    \label{fig:more-localization-cases}
\end{figure*}

Figure~\ref{fig:more-localization-cases} presents six examples of task-conditioned localization. Affordance questions return an interaction point, target-area questions return a set of placement-region points, and trajectory questions return an ordered sequence of image-space coordinates.

\subsection{Agentic Rollout}

\begin{figure*}[p]
    \centering
    \includegraphics[width=\textwidth,height=0.88\textheight,keepaspectratio]{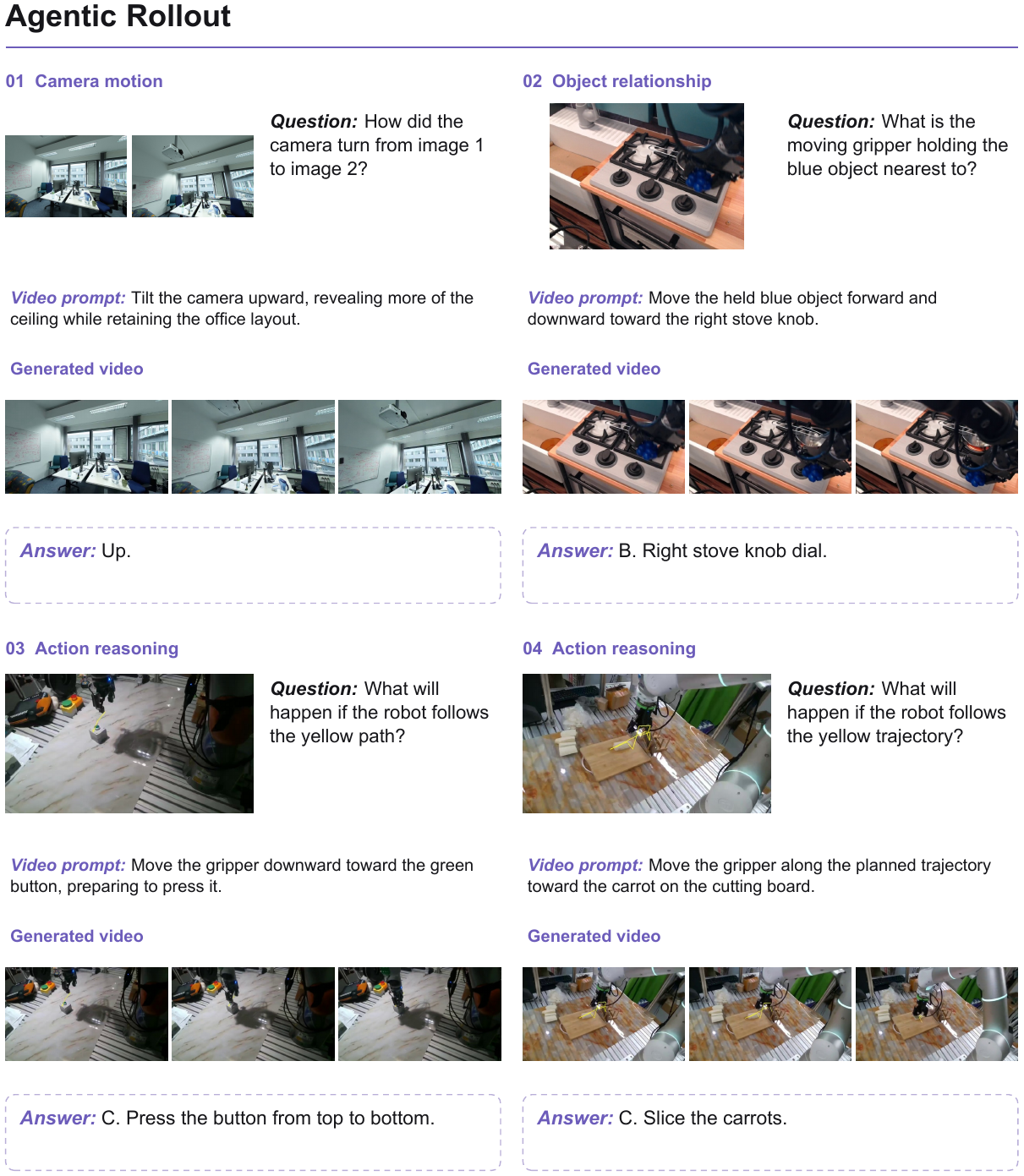}
    \caption{\textbf{Additional agentic rollout examples.} Four cases address camera motion, object relationships, and the anticipated effects of button pressing and slicing. Each case includes the input observation or observations, question, video-generation prompt, sampled generated frames, and answer.}
    \label{fig:more-agentic-cases}
\end{figure*}

Figure~\ref{fig:more-agentic-cases} provides four examples of generation-assisted reasoning about camera motion, object relationships, and action outcomes. Each example connects the input question with a task-specific video prompt, intermediate generated frames, and the final answer.

\subsection{Long-Horizon Visual Planning}

\begin{figure*}[p]
    \centering
    \includegraphics[width=\textwidth,height=0.88\textheight,keepaspectratio]{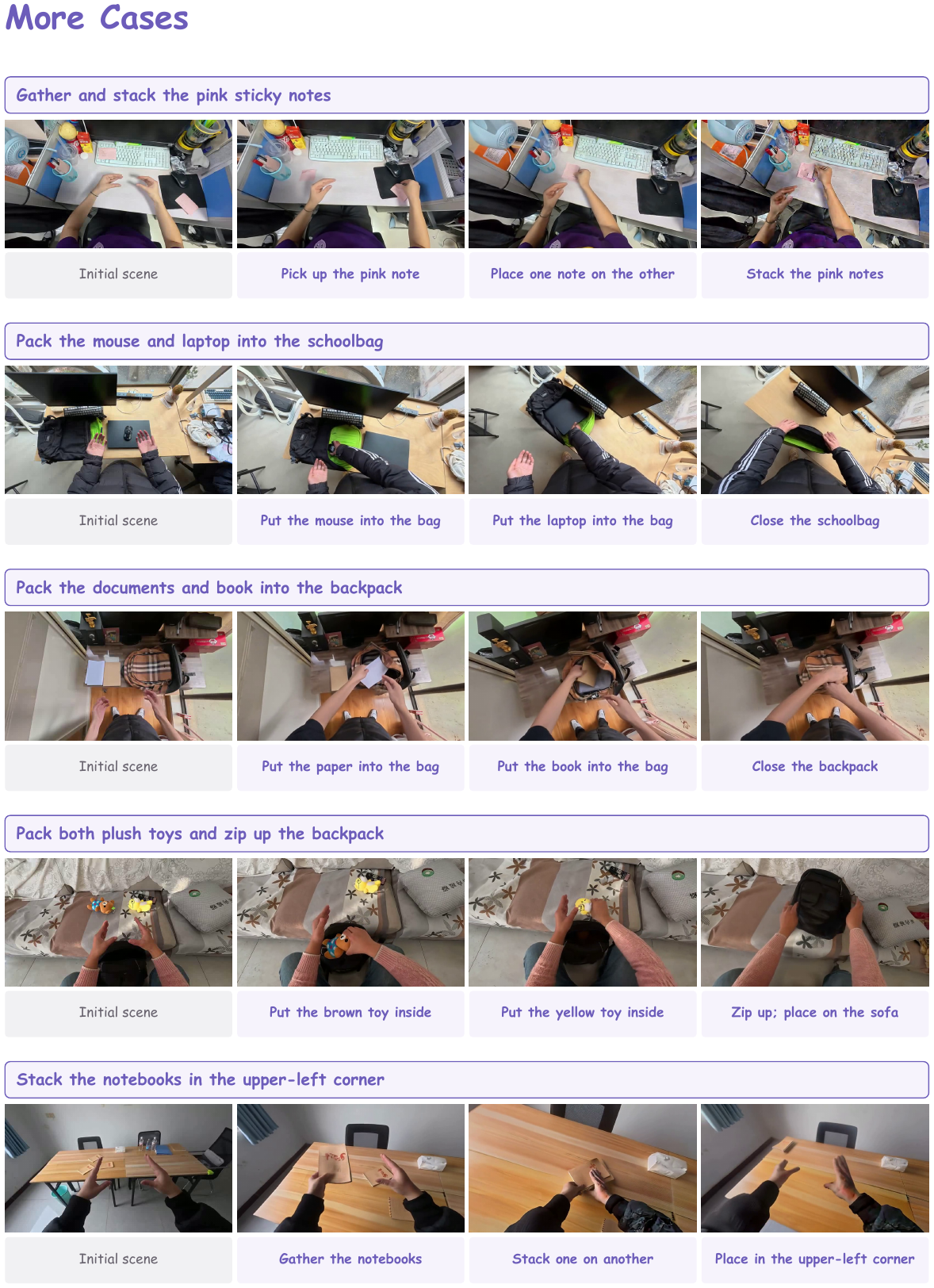}
    \caption{\textbf{Additional long-horizon visual planning examples on Rynn-VP.}}
    \label{fig:more-vp-cases}
\end{figure*}

Figure~\ref{fig:more-vp-cases} provides additional qualitative cases of long-horizon visual planning on Rynn-VP, complementing the examples in the main paper.

\clearpage

\section{Prompt Details}

\subsection{Evaluation and Inference Prompts}
\label{sec:prompt_eval}

% [original]
% Table~\ref{tab:prompts} presents the evaluation prompts and interaction formats used in our experiments. All prompts are presented in a unified Python f-string format.

% [revised]
Table~\ref{tab:prompts} presents the evaluation prompts and interaction formats used in our experiments. Benchmark prompts use Python f-string notation where applicable; planning and agentic interactions are shown as input and output formats.

{
\small
\renewcommand{\arraystretch}{1.3}
\begin{xltabular}{\linewidth}{@{}l >{\raggedright\arraybackslash}X@{}}
\caption{Evaluation prompts and interaction formats.}
\label{tab:prompts}\\
\toprule
\textbf{Benchmark} & \textbf{Prompt Template} \\
\midrule
\endfirsthead
\multicolumn{2}{c}{\tablename~\thetable{} -- continued from previous page}\\
\toprule
\textbf{Benchmark} & \textbf{Prompt Template (Continued)} \\
\midrule
\endhead
\midrule
\multicolumn{2}{r}{Continued on next page}\\
\endfoot
\bottomrule
\endlastfoot

VSI-Bench
& 1.\enspace\texttt{f"\{question\}\textbackslash nAnswer with the option's letter from the given choices directly."}
\newline
2.\enspace\texttt{f"\{question\}\textbackslash n\textbackslash nAnswer the question with an exact number, which should be accurate to at most two decimal places."}
\\
\midrule
MMSI-Bench
& \texttt{f"\{question\}"}
\\
\midrule
MindCube
& \texttt{f"Select the best answer to the following multiple-choice question based on the image.\textbackslash n\{question\}\textbackslash nAnswer with the option's letter from the given choices directly and only give the best option. The best answer is: "}
\\
\midrule
ERQA
& \texttt{f"\{question\}\textbackslash nAnswer with the option letter from the given choices directly."}
\\
\midrule
RoboSpatial
& 1.\enspace\texttt{f"\{question\}. Pinpoint several points within the space. Your answer should be formatted as a list of tuples, i.e. [(x1, y1), ...], where each tuple contains the x and y coordinates... indicating the normalized pixel locations of the points."}
\newline
2.\enspace\texttt{f"\{question\} Answer yes or no."}
\\
\midrule
% Source: RynnScale, commit 0f1365d2f470aa04e739efb7505deca4465267b6, benchmarks/emb_spatial.py.
EmbSpatial
& \texttt{f"Select the best answer to the following multiple-choice question based on the image.\textbackslash nQuestion: \{question\}\textbackslash nOptions:\textbackslash n\{option\_string\}Answer with the option's letter from the given choices directly and only give the best option.The best answer is:"}
\\
\midrule
RefSpatial-Bench
& \texttt{f"Locate \{object\_name\} in this image. Output the point coordinates in JSON format."}
\\
\midrule
\makecell[tl]{ShareRobot\\Affordance}
& \texttt{f"\{question\}The coordinates should be between 0 and 1000, indicating the normalized pixel locations of the point."}
\\
\midrule
\makecell[tl]{ShareRobot\\Trajectory}
& \texttt{f"\{question\}Your answer should be formatted as a list of tuples, i.e. [(x1, y1), (x2, y2), ...], where each tuple contains the x and y coordinates of a point. The coordinates should be between 0 and 1000, indicating the normalized pixel locations of the point."}
\\
\midrule
% Source: RynnScale, commit 0f1365d2f470aa04e739efb7505deca4465267b6, benchmarks/where2place.py.
% question is meta["question"] after load_data removes the coordinate-normalization suffix.
Where2Place
& \texttt{f"\{question\}"}
\\
\midrule
Rynn-VP
&
\raggedright\arraybackslash\hyphenpenalty=10000\relax
\textbf{Input:}\newline
\texttt{[HIGH-LEVEL TASK] \{task\_instruction\}\textbackslash n\allowbreak{}
[INITIAL STATE] \{initial\_observation\}\textbackslash n\allowbreak{}
[PREVIOUS STEPS] \{subtask\_history\}\textbackslash n\allowbreak{}
[CURRENT OBSERVATION] \{current\_observation\}}
\par\smallskip
\textbf{Output format:}\newline
\texttt{<subplan>\{next\_executable\_subtask\}</subplan>\textbackslash n\allowbreak{}
<world\_modeling>\textbackslash n\allowbreak{}
\{scene\_context\_and\_expected\_visual\_evolution\}\textbackslash n\allowbreak{}
</world\_modeling>\textbackslash n\allowbreak{}
<duration>\{duration\_in\_seconds\}</duration>}
\\
\midrule

\makecell[tl]{Rynn-GEU\\Direct}
&
\raggedright\arraybackslash\hyphenpenalty=10000\relax
\texttt{You are a visual reasoning expert. Answer the question
from the supplied visual input. Put only the final answer
inside an <answer> </answer> block.}
\\
\midrule

\makecell[tl]{Rynn-GEU\\Agentic rollout}
&
\raggedright\arraybackslash\hyphenpenalty=10000\relax
\textbf{Input:}\newline
\texttt{[VISUAL INPUT] \{images\_or\_video\_frames\}\textbackslash n\allowbreak{}
[QUESTION] \{question\}}
\par\smallskip
\textbf{Generation request:}\newline
\texttt{<world\_modeling>\textbackslash n\allowbreak{}
\{scene\_context\_and\_requested\_visual\_prediction\}\textbackslash n\allowbreak{}
</world\_modeling>\textbackslash n\allowbreak{}
<duration>\{duration\_in\_seconds\}</duration>}
\par\smallskip
\textbf{Returned visual context:}\newline
\texttt{[VIDEO] \{generated\_video\_frames\}}
\par\smallskip
\textbf{Final response:}\newline
\texttt{<thinking>\{reasoning\}</thinking>\textbackslash n\allowbreak{}
<answer>\{final\_answer\}</answer>}
\\
\end{xltabular}
}

\subsection{Evaluation Hyperparameters}
\label{sec:hyper}

\paragraph{Decoding and preprocessing.}
Unless otherwise specified, sampling is disabled during autoregressive text generation for RynnUnify. For ERQA, RoboSpatial, and ShareRobot, we enable sampling with a temperature of 0.2, \texttt{top\_p} of 0.95, and \texttt{top\_k} of 50.
For image inputs, \texttt{min\_pixels} and \texttt{max\_pixels} are set to $16 \times 32 \times 32$ and $16384 \times 32 \times 32$, respectively. For RefSpatial-Bench and ShareRobot, we increase \texttt{min\_pixels} to $1024 \times 32 \times 32$ to retain finer spatial detail for precise localization.
Input videos are sampled at 2 fps, with a maximum of 512 frames per video. If this limit is exceeded, frames are uniformly subsampled to 512. The minimum pixel budget per frame is $16 \times 32 \times 32$, and the total pixel budget per video is capped at $24576 \times 32 \times 32$. These settings apply to video inputs to the reasoner.

% [original]
% \paragraph{Video generation.}
% On Rynn-VP, RynnUnify generates videos at 480p and 24 fps using 50 diffusion steps, a classifier-free guidance scale of 5, and a fixed random seed. Each segment's duration is specified by the reasoner-predicted \texttt{<duration>} field. Video-generation baselines use a default duration of 4 seconds per step. Rollouts terminate when the model indicates task completion or reaches 15 steps. Results are averaged over three evaluation runs.
% For GEU agentic reasoning, videos are generated at 480p and 24 fps using 35 diffusion steps, a classifier-free guidance scale of 5, and a fixed random seed. The generation mode depends on the number of input images: \texttt{i2v} for a single image and \texttt{fl2v} for multiple images. The generated video's duration is specified by the \texttt{<duration>} field in the generation request.

% [revised]
\paragraph{Video generation.}
On Rynn-VP, RynnUnify generates videos at 480p and 24 fps using 50 diffusion steps, a classifier-free guidance scale of 5, and a fixed random seed. Each segment's duration is specified by the reasoner-predicted \texttt{<duration>} field. Video-generation baselines use a default duration of 4 seconds per step. Rollouts terminate when the model indicates task completion or reaches 15 steps.
For GEU agentic reasoning, videos are generated at 480p and 24 fps using 35 diffusion steps, a classifier-free guidance scale of 5, and a fixed random seed. The generation mode depends on the number of input images: \texttt{i2v} for a single image and \texttt{fl2v} for multiple images. The generated video's duration is specified by the \texttt{<duration>} field in the generation request.

% [original]
% \subsection{Rynn-VP Evaluation}
% \label{app:vp_evaluation}
%
% \subsubsection{Manual Review}
% \label{app:vp_manual_review}
%
% Manual review examines whether the automated Rynn-VP evaluator grounds its judgments in the available visual evidence.
% The review covers evidence interpretation, hallucination assessment, and scoring decisions:
% \begin{itemize}
%     \item \textbf{Evidence interpretation.} Check whether the generated visual sequence supports the evaluator's judgments of subtask completion, task progress, and operation dependencies. Reference demonstrations and timestamped annotations specify the goals and required process evidence, while alternative valid execution sequences remain acceptable.
%     \item \textbf{Hallucination assessment.} Examine judgments concerning object appearance or disappearance, contradictory state changes, physical interactions, and changes in scene layout. Distinguish clear visual inconsistencies from occlusion and insufficient evidence, so that unsupported inferences do not determine the assessment.
%     \item \textbf{Scoring decisions.} Check whether the planning and visual realization judgments follow from the cited evidence and whether the assigned scores are consistent with the scoring criteria, including credit for valid partial progress.
% \end{itemize}
% This review checks the grounding and consistency of the evaluation procedure, complementing the deterministic aggregation of its structured scores.

% [revised]
\subsection{SFT Training Hyperparameters}
\label{app:training_hyperparameters}
Table~\ref{tab:training_hyperparameters} summarizes the optimization settings for reasoner fine-tuning, generator fine-tuning, and WAM post-training.

\begin{table}[htbp]
    \centering
    \caption{SFT training hyperparameters. WAM Stages I--III train the reasoner, generator, and action head, respectively.}
    \label{tab:training_hyperparameters}
    \begingroup
    \small
    \setlength{\tabcolsep}{5pt}
    \renewcommand{\arraystretch}{1.15}
    \begin{tabularx}{\textwidth}{@{}l*{2}{>{\hsize=1.15\hsize\linewidth=\hsize\centering\arraybackslash}X}*{3}{>{\hsize=0.9\hsize\linewidth=\hsize\centering\arraybackslash}X}@{}}
        \toprule
        \multirow{2}{*}{\textbf{Parameter}}
        & \multirow{2}{*}{\makecell{\textbf{Reasoner}\\\textbf{SFT}}}
        & \multirow{2}{*}{\makecell{\textbf{Generator}\\\textbf{SFT}}}
        & \multicolumn{3}{c}{\textbf{WAM Post-training}} \\
        \cmidrule(lr){4-6}
        & & & \textbf{Stage I} & \textbf{Stage II} & \textbf{Stage III} \\
        \midrule
        Optimizer & AdamW & AdamW & AdamW & AdamW & AdamW \\
        Learning rate & $2\times10^{-6}$ & $2\times10^{-5}$ & $1\times10^{-6}$ & $2\times10^{-5}$ &  $1\times10^{-5}$\\
        Global batch size & 1024 & 128 & 1024 & 256 & 256 \\
        Warmup ratio & 0.03 & 0.1 & 0.05 & 0.1 & 0.1 \\
        Gradient clipping norm & 1.0 & 0.3 & 1.0 & 1.0 & 1.0 \\
        \bottomrule
    \end{tabularx}
    \endgroup
\end{table}

\section{Detailed Evaluation Protocol for Rynn-VP}
\label{app:vp_evaluation}

\paragraph{Evaluation setup.}
Rynn-VP evaluates long-horizon visual planning on 200 task instances across six indoor scene categories. Given a natural-language instruction and an initial observation, each system produces a predicted video or an ordered sequence of step images. Reference demonstrations and timestamped annotations are reserved for evaluation. We assess whether the predicted sequence supports an executable plan, faithfully realizes task-relevant actions and effects, and maintains sufficient visual quality.

\paragraph{Task specification.}
Before evaluating candidates, we compile each task into a shared specification containing the original entities, required goal states, necessary action dependencies, protected scene relationships, and allowed variations. Task requirements include object identity, quantity, required transformations, destination, and necessary support or release conditions. The demonstration serves as a solvability witness: alternative strategies, motions, and orders of independent operations are accepted when they satisfy the instruction. The same specification is reused across compared systems.

\paragraph{Evaluation procedure.}
Our evaluator uses GPT-6-sol with the Codex harness and a dedicated evaluation skill defining the evidence requirements and scoring rules. Evaluation comprises five logical stages: observation of objects, actions, and state changes; assessment of visual realization and quality; dense auditing of entity continuity and hallucinations; assessment of planning intelligence; and critical review of the preceding judgments. The standard video evidence package contains at most 50 source frames, including the initial observation, up to four reference anchors, and candidate frames covering the sequence, segment boundaries, and terminal state. Dense auditing uses 30 candidate panels organized into five storyboards. Unresolved contradictions are reviewed before scores are finalized.

\paragraph{Scoring criteria.}
% [original]
% We report Planning Intelligence (PI), Visual Realization (VR), and Visual Quality (VQ), each on a 0--100 scale. PI evaluates the executable task structure supported by the sequence, while VR evaluates its visual execution. Table~\ref{tab:vp_evaluation_criteria} defines their component criteria and within-axis weights. Each axis is computed as the weighted average of its evidence-calibrated component scores. The overall score is

% [revised]
We report Planning Intelligence (PI), Visual Realization (VR), and Visual Quality (VQ), each on a 0--100 scale. PI evaluates the executable task structure supported by the sequence, while VR evaluates its visual execution. Table~\ref{tab:vp_evaluation_criteria} defines their component criteria and within-axis weights. PI and VR are computed as weighted averages of their evidence-calibrated component scores. The overall score is
\begin{equation}
    \mathrm{Overall}
    =
    0.50\,\mathrm{PI}
    + 0.40\,\mathrm{VR}
    + 0.10\,\mathrm{VQ}.
\end{equation}

\begin{table}[t]
\centering
\small
\renewcommand{\arraystretch}{1.15}
\caption{Component criteria for Planning Intelligence (PI) and Visual Realization (VR). Weights are specified within each axis.}
\label{tab:vp_evaluation_criteria}
\begin{tabularx}{\textwidth}{@{}p{0.29\textwidth}Xc@{}}
\toprule
\textbf{Criterion} & \textbf{Definition} & \textbf{Weight} \\
\midrule
\multicolumn{3}{@{}l}{\textbf{Planning Intelligence (PI)}} \\
\midrule
PI-1: Executable Goal Achievement
& Whether the required goals are achieved through executable actions involving the correct original entities.
& 50\% \\

PI-2: Step Validity Rate
& Whether individual operations preserve entity correspondence and satisfy action preconditions with feasible effects.
& 20\% \\

PI-3: Dependency and Order
& Whether necessary causal and temporal dependencies are respected, while allowing equivalent orders for independent operations.
& 15\% \\

PI-4: Plan Completeness
& Whether the sequence covers the necessary operations and establishes durable progress, rather than merely repeating attempts.
& 15\% \\
\midrule
\multicolumn{3}{@{}l}{\textbf{Visual Realization (VR)}} \\
\midrule
VR-1: Visible Goal Achievement
& Whether required object identities, quantities, states, spatial relationships, and necessary support or release are visibly established.
& 20\% \\

VR-2: Action Evidence Fidelity
& Whether relevant contact, grasping, transport, manipulation, and release are supported by clear visual evidence.
& 20\% \\

VR-3: Entity Consistency
& Whether original objects retain consistent identities and quantities, without unexplained creation, duplication, disappearance, or replacement.
& 15\% \\

VR-4: Effect and State Continuity
& Whether task-relevant changes have plausible causes, persist when required, and are inherited by subsequent operations.
& 15\% \\

VR-5: Cross-Clip Coherence
& Whether object identities, scene layouts, container states, and ongoing interactions remain consistent across generated segments.
& 10\% \\

VR-6: Physical Feasibility
& Whether depicted actions and configurations are possible given contact, accessibility, support, and containment constraints.
& 10\% \\

VR-7: Physical Plausibility
& Whether motion and deformation exhibit credible responses to gravity, contact, and support.
& 7\% \\

VR-8: Perceptual Fluency and Quality
& Whether visual execution is clear and temporally readable, with limited blur, flicker, ghosting, tearing, or motion discontinuities.
& 3\% \\
\bottomrule
\end{tabularx}
\end{table}

\paragraph{Visual quality.}
VQ provides a holistic assessment supported by seven diagnostic dimensions: image clarity, task observability, freedom from artifacts, motion readability, temporal stability, freedom from unintended freezing, and camera continuity. These dimensions explain the quality judgment rather than define a fixed arithmetic average. VQ is assessed separately from the low-weight perceptual component VR-8. For ordered step images, quality is judged according to the native image format; the absence of continuous motion is not itself penalized.

\paragraph{Hallucinations and evidence gaps.}
We distinguish physical execution violations, entity lifecycle violations, and scene consistency violations. These include motion without a plausible cause, unexplained object duplication or replacement, and unsupported changes to containers or scene structure. Confirmation requires examining the affected original entity, its before-and-after states, the action responsible for the change, and plausible alternative explanations. Missing contact in sparsely sampled or occluded evidence is recorded as an evidence gap unless additional observations support a violation. Ordered step images follow the same task and entity criteria, while unshown intermediate actions remain unverified.

\paragraph{Goal verification and partial credit.}
Complete execution requires sufficient support for all necessary goal conditions. For placement tasks, we check the specified destination and appropriate support or release; continued hand contact alone does not imply that placement is unfinished. For folding tasks, we inspect persistent folds and reduced spatial extent while allowing different valid folding strategies. Valid subgoals retain partial credit when the full task remains incomplete. Repeated attempts do not accumulate additional credit without new progress, and later actions that undo a required state reduce terminal goal achievement. Apparent completion based on a substituted object or an invalid task-critical state transition does not establish completion of the original task.

\paragraph{Human-guided calibration.}
Ten participants, all holding a master's degree or higher, reviewed evaluator judgments across the 200 benchmark tasks. Each participant reviewed 30 cases in rotating batches, yielding 300 participant--case review assignments with overlapping coverage. Review focused on task progress, terminal goal satisfaction, hallucination detection and severity, and appropriate partial credit. Their feedback guided iterative refinement of the evaluation skill and review of affected judgments until the resulting assessments aligned with human preferences regarding the correctness and quality of visual task execution.

\paragraph{Reporting.}
We report mean PI, VR, VQ, and Overall scores across evaluated tasks, using unrounded per-task values for aggregation. Reports retain criterion-level judgments, supporting visual evidence, and unresolved uncertainties. Missing or invalid candidates are marked unevaluable, with evaluation coverage reported explicitly.

\section{Detailed Training Protocol of Agentic RL}
\label{app:rl}
\label{app:rl_recipe}

\paragraph{Training settings.}
We train the agent with 22,007 training questions and 883 questions for in-training validation. The agent is initialized from RynnUnify's 8B reasoner, while the visual-generation model, reference policy, and semantic judge remain frozen. Each update contains 64 questions and eight trajectories per question, yielding 512 trajectories. Routable questions use four direct and four generation-assisted trajectories; native multi-image questions use eight direct trajectories. We optimize the agent for one epoch using Adam with a constant learning rate of $10^{-6}$, shuffling the questions and cyclically padding the final batch to obtain 344 updates. Responses are sampled at temperature 1.0 with at most 2,048 generated tokens per reasoner turn. We set the advantage stabilizer to $10^{-6}$, the lower and upper GSPO clipping thresholds to $0.0003$ and $0.0004$, the routing temperature to $0.1$, and the routing, KL, and entropy coefficients to $0.01$, $0.001$, and $0.001$, respectively. Validation uses greedy decoding and autonomous D/G selection every ten updates, and checkpoints are saved every twenty updates.

\paragraph{Interaction protocols.}
A direct trajectory answers from the original observations, whereas a generation-assisted trajectory produces a \texttt{<world\_modeling>} instruction and a \texttt{<duration>} value, receives the generated video, and then outputs its reasoning and final answer in \texttt{<thinking>} and \texttt{<answer>} blocks. Each trajectory invokes the generator at most once, with a requested duration in $(0,10]$ seconds. The generator uses the observed conditioning images; for physical prediction from an observed video prefix, it uses the final decoded frame of that prefix. Reference answers and future videos are excluded from the policy context. The generation request and subsequent answer share one sequence-level importance ratio and advantage, while prompts, forced route prefixes, and visual observations are excluded from the action-token loss. Routing supervision is enabled only when a question is routable, all eight trajectories satisfy their output protocols, and at least one branch has a positive mean reward:
$m(x)=\mathbf{1}[\mathrm{routable}(x)]\prod_{i=1}^{8}F(\tau_i)\mathbf{1}[\max(\mu_D,\mu_G)>0]$.
The masked routing loss is normalized by the full question batch, and its soft targets are detached from the gradient. Questions with a zero routing mask retain their within-branch policy objective.

\paragraph{Reward design.}
\label{app:rl_rewards}
Both branches use the same task-specific answer score, ensuring that their reward difference measures the benefit of generation under a shared criterion. For routable questions, the final reward is
$r(\tau)=F(\tau)A_k(\widehat{y},y^\star)$,
where $F(\tau)$ is a binary protocol-validity gate. Native multi-image questions use their native answer score directly. Numerical and geometric tasks use the common shaping function
\begin{equation}
    C(e)=\left[\max\left(0,1-\frac{e}{0.30}\right)\right]^3,
    \qquad
    A_k=C(e_k).
    \label{eq:app_rl_reward_shaping}
\end{equation}
Table~\ref{tab:app_rl_reward_details} specifies the task-dependent errors and discrete scores.

\begin{table}[t]
    \centering
    \small
    \renewcommand{\arraystretch}{1.2}
    \caption{Task-specific answer rewards. Numerical predictions and references are denoted by $\widehat{a}$ and $a^\star$; predicted and reference point sets or trajectories are denoted by $P$ and $P^\star$.}
    \label{tab:app_rl_reward_details}
    \begin{tabularx}{\textwidth}{@{}p{0.23\textwidth}X@{}}
        \toprule
        \textbf{Task} & \textbf{Scoring rule} \\
        \midrule
        Multiple choice
        & Exact match between the extracted prediction and reference option. \\

        Physical quantities and spatial distance
        & $A_k=C(e_k)$, with
        $e_k=|\widehat{a}-a^\star|/|a^\star|$
        after compatible-unit conversion. \\

        Affordance and area
        & $A_k=C(e_k)$, with
        $e_k=\frac{1}{2000}\left[
        \frac{1}{|P|}\sum_{p\in P}\min_{q\in P^\star}\|p-q\|_2+
        \frac{1}{|P^\star|}\sum_{q\in P^\star}\min_{p\in P}\|q-p\|_2
        \right]$. \\

        Trajectory
        & $A_k=C(e_k)$, with
        $e_k=d_{\mathrm{DF}}(\widetilde{P},\widetilde{P}^{\star})/L(P^\star)$,
        where $d_{\mathrm{DF}}$ is discrete Fr\'echet distance and
        $L(P^\star)=\sum_{j=2}^{|P^\star|}\|p^\star_j-p^\star_{j-1}\|_2$. \\

        Textual answers
        & Normalized deterministic matching where applicable, followed by binary semantic judgment for unmatched answers. \\

        Mess
        & $A_k=\ell/4$, where $\ell\in\{0,1,2,3,4\}$ measures correctness and coverage of predicted interaction effects. \\
        \bottomrule
    \end{tabularx}
\end{table}

Numerical parsing preserves signs and checks unit compatibility. A zero-valued reference receives full credit only for an exact zero prediction. Affordance and area scoring uses point sets in the $0$--$1000$ image-coordinate system. For trajectory scoring, we parse up to ten ordered points and resample each trajectory to fifteen points at uniform arc-length intervals, producing $\widetilde{P}$ and $\widetilde{P}^{\star}$. The normalization uses the reference arc length before resampling. A zero-length reference receives full credit only when the Fr\'echet distance is zero. Invalid numerical or geometric predictions receive zero answer reward.

Semantic judgments use the frozen Qwen3.8-27B-FP8 judge, which assesses correctness and completeness from the question, reference answer, and extracted final answer, excluding the agent's reasoning trace and generation request. Mess judgments assess effects on the main non-manipulated objects, including motion direction and magnitude, rotation, and support or contact changes. Levels $0$--$4$ indicate no substantive correct effects, a small minority of correct effects, substantial partial correctness, mostly correct effects with limited omissions, and complete correctness, respectively. Malformed or incomplete judgments trigger up to three evaluation attempts. Generator failures are treated as missing feedback rather than zero rewards, distinguishing infrastructure failures from incorrect predictions.